\documentclass{article}
\usepackage[utf8]{inputenc}
\usepackage{hyperref}
\usepackage{amsmath,amsfonts,graphicx,dsfont,float,ntheorem,physics,multirow,listings}
\usepackage[ruled,vlined]{algorithm2e}
\DeclareMathOperator*{\argmax}{arg\,max}
\usepackage{booktabs} 
\usepackage{bm, bbm}
\usepackage{amssymb}

\begin{document}

\title{A Reinforcement Learning Based Approach to Play Calling in Football}
\author{Preston Biro and Stephen G. Walker}

\maketitle

\begin{abstract}
With the vast amount of data collected on football and the growth of computing abilities, many games involving decision choices can be optimized. The underlying rule is the maximization of an expected utility of outcomes and the law of large numbers. The data available allows us to compute with high accuracy the probabilities of outcomes of decisions and the well defined points system in the game allows us to have the necessary terminal utilities. With some well established theory we can then optimize choices at a single play level.
\end{abstract}


\noindent \textbf{Corresponding Author: Preston Biro,} University of Texas at Austin, Department of Statistics and Data Sciences, Austin, TX 78712-1823, USA, e-mail: prestonbiro@utexas.edu\\
\noindent \textbf{Stephen G. Walker} University of Texas at Austin, Department of Mathematics, Austin, TX 78712-1823, USA, e-mail: s.g.walker@math.utexas.edu

\section{Introduction}

With the advances in computer power and the ability to both acquire and store huge quantities of data, so goes the corresponding advance of the machine (aka algorithm) to replace the human as a primary source of decision making.  The number of successful applications is increasing at a rapid pace; in games, such as Chess and Go, medical imaging and diagnosing tumours, to automated driving, and even the selection of candidates for jobs.
The notion of reinforcement learning is one key principle, whereby a game or set of decisions is studied and rewards 
recorded so a machine can learn long term benefits from local decisions, often negotiating a sequence of complex decisions. For example, Silver et al. (2017) discuss how a machine can become an  expert at the game Go simply by playing against itself, with Bai and Jin (2020) looking at more general self--play algorithms. 

The amount of data available within the game of football is now reaching levels from which a comprehensive understanding of the outcomes of decisions can be accurately obtained. If outcomes to decisions are well tabulated then determination of optimal decisions can be made. The theory is well known and sits under the general framework of \emph{Decision Analytics}; in words, making decisions with uncertainty.

Early work on decision making within football has been done by Carter and Machol (1971) who introduced the idea of Expected Points but only within a limited context, such as for first downs. Carroll et al. (1988) used a model approach to expected points and directed attention to the valuation of yards gained, depending on which yards the gains were over. Other specific aspects of decisions with a game, such as sudden--death, have been considered by Sahi and Shubi (1987), and down-and-goal by Boronico and Newbert (2007), and fourth down choices by Yam and Lopez (2019).

To introduce our approach and the ideas at the most simplest of levels, consider a repeated game in which action $a$, from a choice of actions $\mathbb{A}$, is made and for each action the reward assigned, $U(s)$,  for $s\in\mathbb{S}$, is random and occurs with probability $P(s\mid a)$. The optimal action, to maximize the reward over a long sequence of plays, is to select the action $a$ which maximizes the expected reward,
$$\bar{U}(a)=\sum_{s\in \mathbb{S}}U(s)\,P(s\mid a).$$
Optimizing choices is a well studied problem; from neural networks,  Hopfield and Tank (1985), to the bandit problem, Woodroofe (1979), and for an economic perspective, see, for example, Hirschleifer \& Riley (1992).

This problem can be solved if the collection of $P(s\mid a)$ are known and the utilities of each end state $s\in \mathbb{S}$; i.e. $U(s)$, are specified.  The aim in this paper is to demonstrate how the vast quantities of data within football allow us to determine the probabilities of outcomes associated with actions taken on the field and we argue it is straightforward to specify the required utilities since the game is based on a point scoring system.

The game of football is remarkably simple in structure. It is constructed from a sequence of \emph{drives}. A drive involves a starting position for the offensive team, which comprises three key ingredients: first down, yardage to secure another first down, and yardage to the goal line or equivalently the line of scrimmage. The drive consists of a sequence of \emph{plays} and a play can be described with the three pieces of information; which we write as $s=(DOWN,DIST,LOS)$, where $DOWN$ is the number of down, $DOWN\in\{1,2,3,4\}$, $DIST$ is the yardage to another first down, and $LOS$ is the yardage to a touchdown. This is also essentially the description of states described in, for example,  Goldner (2017) and the references mentioned in the paper.

The actions at each play we categorise as two types; $a\in\{R,P\}$, where $R$ denotes ``run'', and $P$ denotes ``pass''. In special cases other actions are also possible, such as a punt or field goal attempt. However, these are typically actions taken when no other realistic option is available. A drive ends with entry to a terminal state, which is a score; i.e. a touchdown or a field goal or a yield of possession to the other team. Let us label these terminal states to belong to the set $\mathbb{S}_T$.

Hence, a drive consists of a sequence of states $(s_k)$, for $k=1,\ldots,N+1$, with $s_{N+1}\in\mathbb{S}_T$, and $N$ is the number of plays within the drive. The actions made are $(a_k)$, for $k=1,\ldots,N$. The aim, as in the game, and what we will endeavour to provide from a theoretical perspective, is to select the actions at each state of play to maximize the reward; i.e. the score, at the end of the drive. In particular, we can show how to do this when we know the $P(s^*\mid s,a)$, for all $s\in\mathbb{S}$, $s^*\in\mathbb{S}\cup\mathbb{S}_T$ and $a\in\mathbb{A}$, and have $U(s)$, for all $s\in\mathbb{S}_T$. As we have mentioned, the $U(s)_{s\in\mathbb{S}_T}$, are easily specified, and the probabilities can be accurately estimated from the vast swathes of data available.

Alternative approaches to decision making within football include Jordan et al. (2009), and papers cited in this paper. These authors rely on models and online decisions using updating rules. They also account for the attitude of risk of the team. We, on the other hand, do not require models or any underlying assumptions of the team's mindset, since we are able, using machine learning techniques with the huge quantities of data and the specification of elementary utilities, to obtain optimal decisions for every non terminal state.  The upshot is that optimality then relies on the law of large numbers to put the optimal theoretical choices into practice to maximize expected scores.

A decision theoretic paper more along the lines of our own is Goldner (2017). As we do, the author considers the Markov model with transition probabilities obtained empirically. A concern in Goldner (2017) is about the lack of frequencies for some plays and smoothing techniques are used to fill in the gaps. However, our argument is that limited transition frequencies indicates small probabilities. And lack of observations from specific states, though they do exist, indicates such states rarely occur and hence estimating the probabilities of transition with the limited information works since they will contribute insignificantly to the overall analysis. This subject of data information is discussed in detail in Appendix A.

Another aspect of the Goldner (2017) analysis is the mixing of drives; to see the game as a whole and to consider and quantify ideas such as  ``future expected scores if a drive does not end in a score''. Other similar concepts such as the expected points and expected points added have been studied in Burke (2014). Recent work on this and the Win Probability theme is detailed in Yurko et al. (2018). 

We argue to the contrary; that each drive can be analyzed as a self contained unit within a game and the rewards associated with the terminal states and transition probabilities are sufficient to determine optimal decisions. We do discuss modifications to the analysis for optimal decisions if a drive needs to achieve a certain outcome; for example, a number of points in a particular amount of time or, for example, the team needs to run out the clock.   

With our data driven and machine learning techniques we are able to provide optimal actions for each state of play. With these optimal choices we can analyse the performance of teams as a percentage of their actions as they relate to the optimal choices. This yields a highly informative and illuminating figure which fully supports our approach.  

To describe the layout of the paper: in section 2 we provide some further details on the theory on which we rely, and we demonstrate in a sequence of illustrations which become more realistic to the game of football while holding the key principles on which we depend. Section 3 provides the information on how we are able to set all the necessary utilities of states. These combined with the empirically evaluated probabilities give us the ability to evaluate the optimal decisions. Section 4 presents our results and provides a comparison with alternative strategies. We also provide some insights on how teams perform in terms of their percentage of optimal decisions. In this section we also
consider special plays. Finally section 5 contains some conclusions and pointers to future work.

\section{Background}
Football differs from many of the previously discussed applications in that the action space is small. As opposed to a game like Chess or Go where a player usually has on the order of tens or even hundreds of sequences of moves available at any particular state, on most football plays there are only two reasonable actions that can be chosen; run or pass. The exceptions to this generally only come on fourth down or on the last play of a half, where a team may choose to kick a field goal or punt, depending on the score and field position. We can view a drive 
as a progression of states, where a state can be characterized by the current down, current distance needed for a first down, and the current line of scrimmage. As a team progresses through a drive, they will get the option to choose a singular action - run or pass - at which point, we can view the outcome; i.e. the next state, of this choice as a draw from a probability distribution corresponding to the choice of action given the current state.
This continues until the offensive team reaches one of the terminal states of the drive.

In the following illustrations we extract some fundamentals of the game to show how we proceed with deriving optimal plays.
The major difference between our more simple examples and the game itself will lie in the different number of states. However, the principles learnt from these simple cases can be carried forward when the numbers of states becomes larger.   

\subsection{A Five State System}

\begin{figure}[H]
    \includegraphics[width = 2.25in,height=2in]{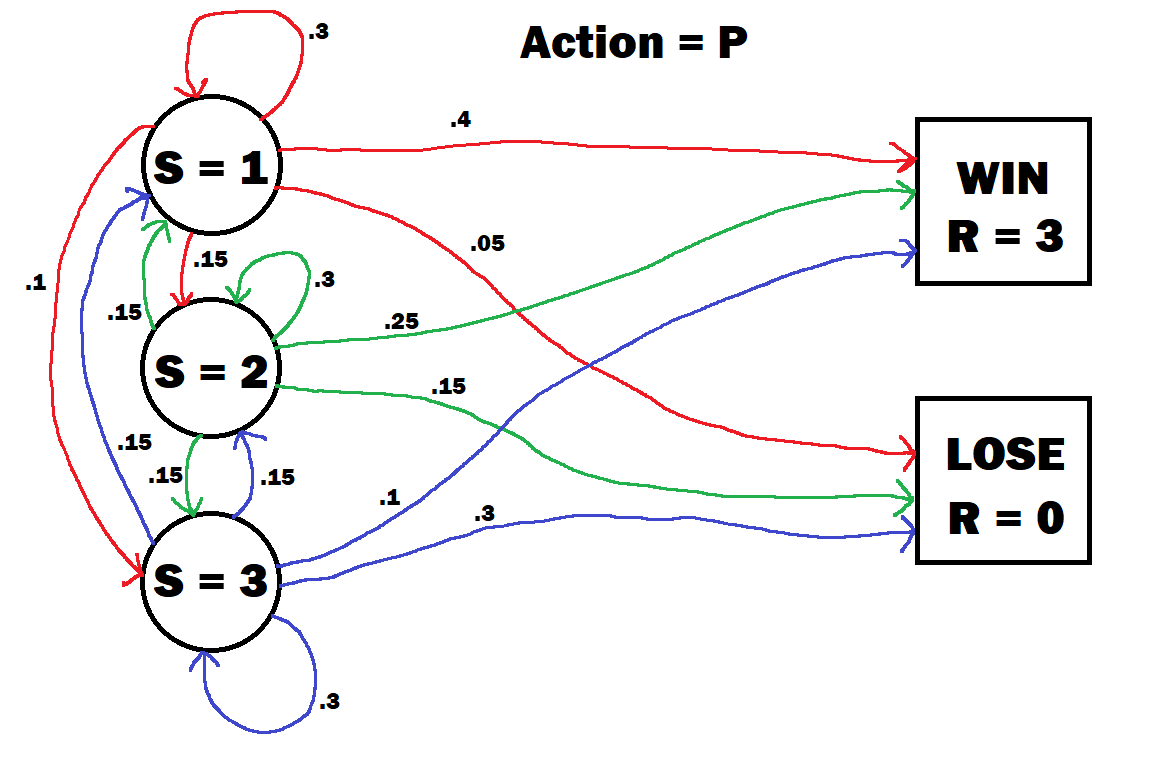}%
    \includegraphics[width = 2.25in,height=2in]{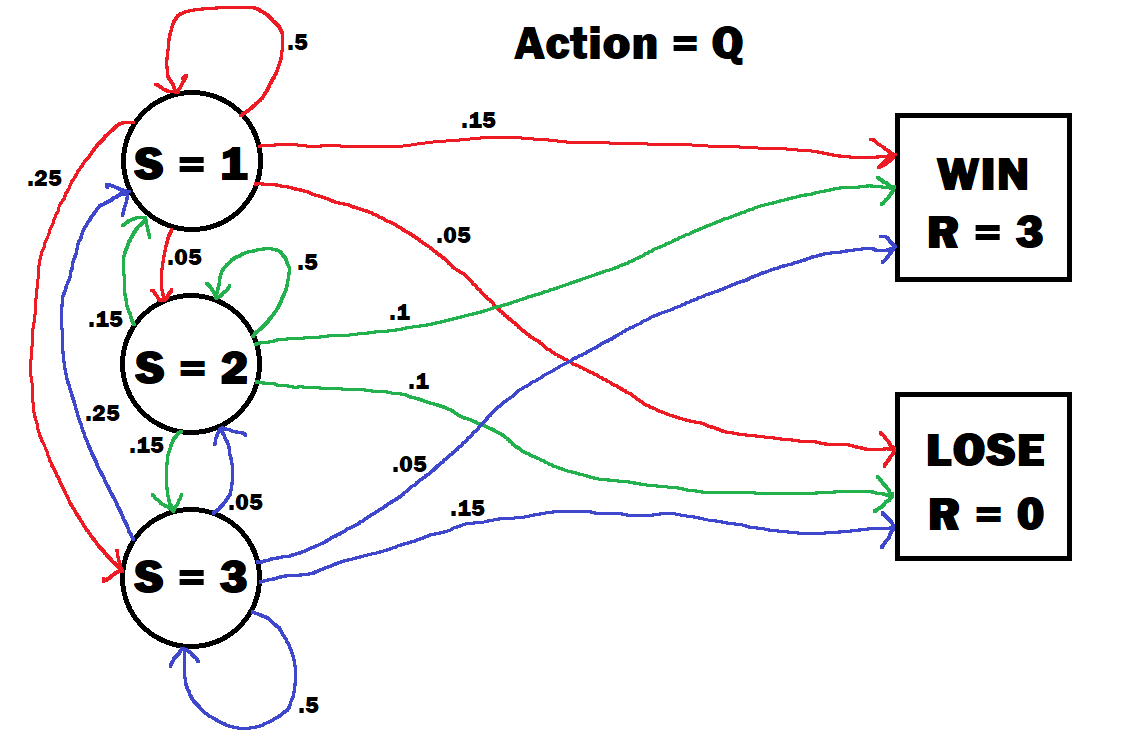}
    \caption{A toy example of a Markov decision process.}
    \label{fig1}
\end{figure}

We start off by considering an analogous game where the number of states is reduced to five. Three states are non--terminal while two are terminal.   So consider the system of states connected via actions depicted in Fig.~\ref{fig1}. There are five states, two of which are terminal ($WIN$ and $LOSE$) with specified rewards, $R(WIN) = 3$, and  $R(LOSE) = 0$. We use $R$ for reward here to keep it more explicit, but for more formal derivations we will use $U$ for utility. The other states, which we will refer to as intermediate states, designated $\mathbb{S}_I$, have no reward from being within the state. However, these states can have a different utility in the sense of being in one rather than another can give rise to a higher expected reward on entering a terminal state.
The states are connected via actions $P$ and $Q$, and have specified transition probabilities. That is, an action, a choice of two, taken in one state would open up a probability vector determining the next state. Hence, the process can be fully determined by two stochastic matrices with $P$ carrying the probabilities for one action and $Q$ for the other.

So, for example, $P$ will be a $3\times 5$ matrix of the form
$$P=\left(\begin{array}{lllll}
p_{11} & p_{12} & p_{13} & p_{14} & p_{15} \\
p_{21} & p_{22} & p_{23} & p_{24} & p_{25} \\
p_{31} & p_{32} & p_{33} & p_{34} & p_{35} \\
0 & 0 & 0 & 1 & 0 \\
0 & 0 & 0 & 0 & 1
\end{array}\right),
$$
where for each row the sum of probabilities is 1, and the first three rows indicate an intermediate state while the last two rows represent the terminal states.


The aim is to find the optimal action at each state so as to maximize the reward at the end of the game.  
Choosing an action requires maximizing the probability that the eventual terminal state is $s = WIN$, as it is the only state that offers a positive reward. However, since the intermediate states are all co--dependent, their utilities or expected rewards, are interconnected. 

We will define the utility of state $s \in \mathbb{S}_I$ by $U(s)$;  the expected reward if optimal actions were taken at each state. 
For terminal states, $s \in \mathbb{S}_T$, the utility  is simply the reward achieved for arriving at the state, so $U(s) = R(s)$. 
Similarly, we will define the value, also known as an elementary utility, of an action $a$ taken at state $s$ as $V(\mathbb{S} = s,A = a) = V(s,a)$. This is the expected final reward for taking action $a$ at state $s$. 
The value of a state/action pair can be thought of as the weighted sum of the utilities of future states, weighted by their respective probabilities of transitioning to each of the future states. These ideas can be expressed mathematically in the following way, and known as the Bellman equations,
\begin{align*}
        V( s, a) &= \sum\limits_{s^\star} U(s^\star) P (s^\star \vert s, a) \\
        U( s) &= \max_{a^\star} V( s,a^\star).
\end{align*}
See Bellman (1957).

We can make a few simplifications due to the structure of our specific setup. Since the state space is split between intermediate and terminal states, $\mathbb{S}_I$ and $\mathbb{S}_T$, we can split up the summation as well, allowing us to have a term that is dependent upon only terminal utilities. 
Additionally, since intermediate states can transition to themselves, the utility of a state is dependent upon itself. By algebraically manipulating the equation, the self-dependence can be removed by dividing the summation of the other states by the probability of not self--transitioning. Thus the equations can be rewritten as follows:

\begin{align*}
  &  V(s, a) 
    = \sum\limits_{s_i^\star \in \mathbb{S}_I} U(s_i^\star) P( s_i^\star \vert s,  a) 
    + \sum\limits_{s_t^\star \in \mathbb{S}_T} U(s_t^\star) P(s_t^\star \vert s,  a)  \\
 &   U(s) 
    = \max_{a^\star} \left[U(s) P(s \vert  s,  a^\star)
    + \sum\limits_{s_i^\star \in \mathbb{S}_I: s_i^\star \neq s } U(s_i^\star) P(s_i^\star \vert s,  a^\star)
    + \sum\limits_{s_t^\star \in \mathbb{S}_T} U(s_t^\star) P( s_t^\star \vert s, a^\star) \right] 
\end{align*}
which implies    
\begin{equation}\label{utility}
U(s) = \max_{a^\star} \frac{\sum\limits_{s^\star \in \mathbb{S}: s^\star \neq s } U(s^\star) P(s^\star \vert  s, a^\star)}{1 - P( s \vert  s, a^\star)}.
\end{equation}
In the case that a state cannot self--transition, $P(s \vert s,  a) = 0$, and then the last line in (\ref{utility}) has the denominator as 1.

This framework allows us to express all the utilities in terms of one another, giving us a set of equations we can jointly maximize. For example, we can calculate the utility of state $s = 1$ as follows;
$$ U(1) = \max\bigg(V(1,P),V(1,Q)\bigg) $$
where

\begin{align*}
    V(1,P) 
        &= 0.3 U(1) + 0.15 U(2) + 0.1 U(3) + 0.4 U(WIN) + 0.05 U(LOSE)\\
         &= 0.3 U(1) + 0.15 U(2) + 0.1 U(3) + 1.2\\
        V(1,Q) &= 0.5 U(1) +0 .05 U(2) + 0.25 U(3) +0 .15 U(WIN) + 0.05 U(LOSE)\\
        &= 0.5 U(1) + 0.05 U(2) + 0.25 U(3) + 0.45
\end{align*}        
 so       
$$      
        U(1) = \max\left\{\frac{0.15 U(2) + 0.1 U(3) + 1.2}{0.7},\frac{0.05 U(2) + 0.25 U(3) + 0.45}{0.5}\right\}.
$$
Similarly, we can write the utilities for the other two states as follows,
\begin{align*}
    U(2) &= \max\left\{\frac{0.15 U(1) + 0.15 U(3) + 0.75}{0.7},\frac{0.15 U(1) + 0.15 U(3) + 0.3}{0.5}\right\}\\
        U(3) &= \max\left\{\frac{0.15 U(1) + 0.15 U(2) + 0.3}{0.7},\frac{0.25 U(1) + 0.05 U(2) + 0.15}{0.5}\right\}
\end{align*}

\begin{algorithm}[H]
        \SetAlgoLined
        \KwResult{A deterministic policy, $\pi$, that designates the optimal action at each intermediate state }
         Choose a small threshold $\theta > 0$, determining accuracy of estimation\;
         Initialize $U(s)$ for all $s \in \mathbb{S}_I$ arbitrarily\;
         Initialize $U(s)$ for all $s \in \mathbb{S}_T$ as designated terminal rewards\;
         Initialize $\Delta > \theta$, arbitrarily;\\
         \While{$\Delta > \theta$}{
          \For{each $s \in \mathbb{S}_I$}{
            $u \gets U(s)$\\
            $U(s) \gets \max\limits_{a^\star} \frac{\sum\limits_{s^\star \in \mathbb{S}: s^\star \neq s } U(s^\star) P( s^\star \vert s, a^\star)}{1 - P(s \vert s, a^\star)}$\\
            $\Delta \gets \Delta + (U(s) - u) ^ 2$\\
          }
         }
Return $\pi$ such that for all $s \in \mathbb{S}_I$\\
         $\pi(s) = \argmax\limits_{a^\star} \frac{\sum\limits_{s^\star \in \mathbb{S}: s^\star \neq s } U(s^\star) P( s^\star \vert  s,  a^\star)}{1 - P( s \vert  s, a^\star)}$
         
     \caption{Value Iteration, Sutton and Barto (2018)}
\end{algorithm}

Solving these three equations requires a joint maximization, and thus will need an algorithm to do this.  
The following algorithm is a modified version of Value Iteration, appearing in Sutton and Barto (2018), and incorporating the structure exhibited by this example to produce faster convergence.
This algorithm can be seen as initializing the utilities at the start of the program, sequentially updating each of the values using their specified equations, and iterating until a convergence threshold is reached. The output of the program is an action recommendation at each state, which we will refer to as the policy, $\pi$. This can be considered a \textit{greedy} policy, as each of the action recommendations $\pi(s)$ give that play which maximizes the expected reward. 

\begin{align*}
    U(1) = 2.37,\hspace{2mm}&\hspace{2mm} \pi(1) =  P\\
    U(2) = 1.94,\hspace{2mm}&\hspace{2mm} \pi(2) =  P\\
    U(3) = 1.68,\hspace{2mm}&\hspace{2mm} \pi(3) =  Q
\end{align*}

\begin{figure}[H]
        \includegraphics[width = 2.25in,height=2in]{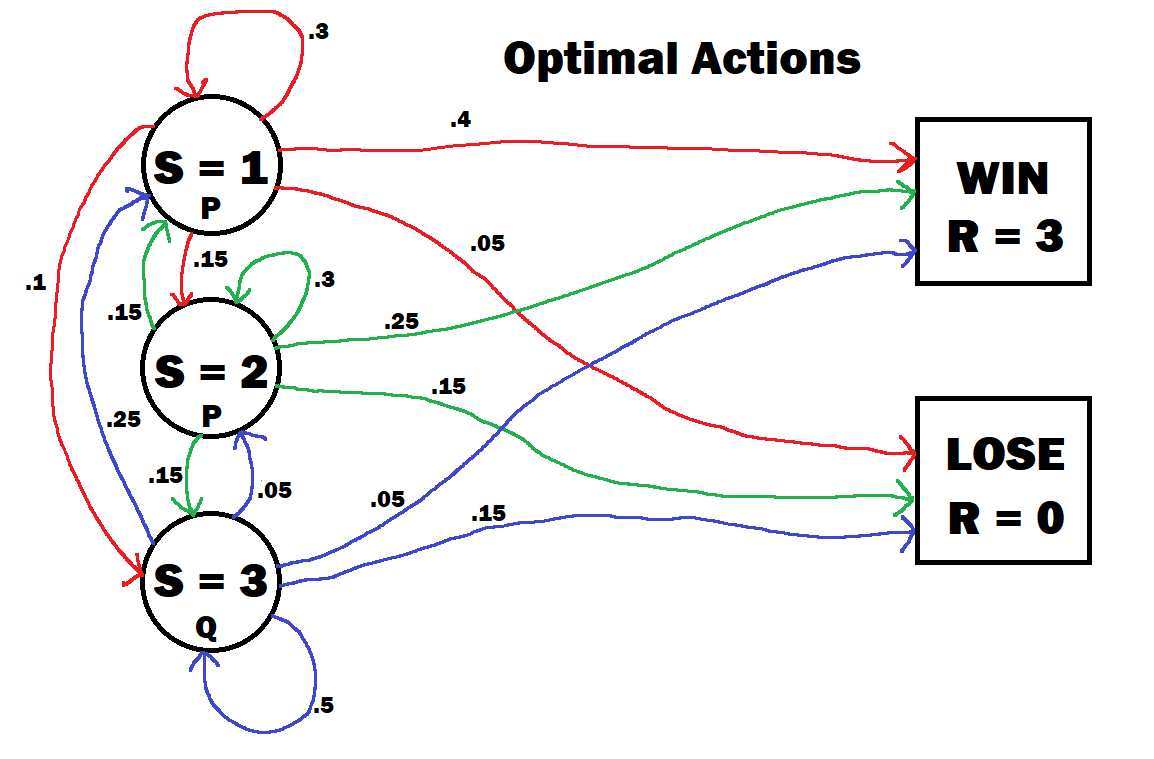}%
        \includegraphics[width = 2.25in,height=2in]{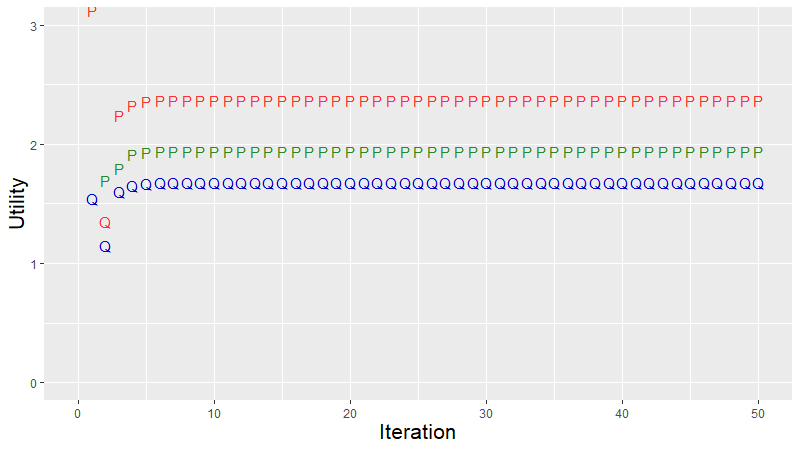}
        \caption{The toy example solved, showing optimal actions converging with discovered intermediate utilities.}
        \label{fig2}
\end{figure}

Applying the algorithm to our example allows us to find the utilities and optimal policy. As the utility plot in Fig.~\ref{fig2} shows, convergence is 
rapid. Even with utilities starting far from their final values, the algorithm still converges in about five iterations.

\subsection{An Ordered System}
\begin{figure}[H]
        \makebox[\textwidth][c]{\includegraphics[width = 4.5in,height=3in]{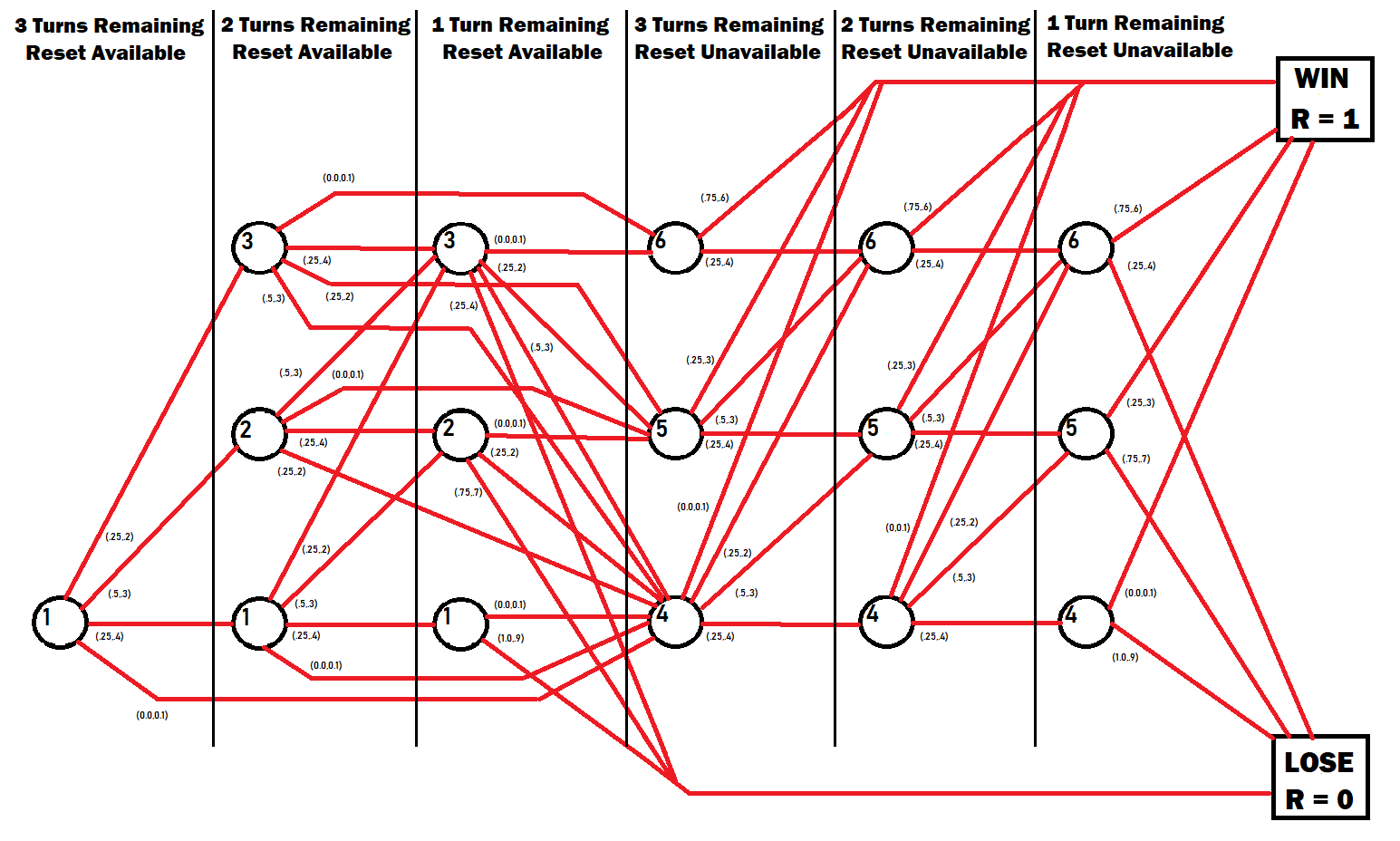}}
        \caption{An ordered toy example of a Markov decision process.}
        \label{fig3}
\end{figure}

Now consider a new game displayed in Fig.~\ref{fig3} which is becoming closer to the game of football. The states are again split between intermediate and terminal, with two terminal states  $WIN$ and $LOSE$, with corresponding utilities  $U(WIN) = 1$ and $U(LOSE) = 0$. The goal is to progress to the $WIN$ state by moving forward in position. There are six intermediate positions, labeled one through six, at which the player can choose action $P$ or $Q$ and progress forward either zero, one, two, or three positions. The amount of positions the player moves forward is probabilistically driven by their action choice. The player gets three turns to reach at least position four, at which point they will have three turns to complete the game. If the player ever runs out of turns, they lose. In the following figure, the transition probabilities are shown in parentheses as $P( s \vert  s, P)$ and $P (s \vert  s,  Q)$.

This game's structure has characteristics that differentiate it from the previous example. Here, the state space requires a much more robust characterization, as it now requires information about the current position, the amount of turns remaining, and the availability of a turn reset (whether or not state four has been reached). Also in this setup, there is a clear ``downhill" nature that can be observed in the transitioning between states. At no point can a state self--transition, nor can it ever transition to a previous state, implying the game must move in one direction. We can think of a game with this structure as an \textit{Ordered Markov Decision Process}. It is also clear that this game has a limit in how long it will last, with a maximum of six actions taken over the course of the game. 

This new characterization gives some helpful properties that will allow for a speedier convergence in utility calculations. The following algorithm utilizes these updates to allow for one-step optimization of utilities.

\begin{algorithm}[H]
        \SetAlgoLined
        \KwResult{A deterministic policy, $\pi$, that designates the optimal action at each intermediate state }
         Initialize $U(s)$ for all $s \in \mathbb{S}_T$ as designated terminal rewards\;
         
         \textbf{Reorder $\mathbb{S}_I$ such that $s_j$ depends only on $s_1$, \dots $s_{j-1}$ and $\mathbb{S}_T$}
    
          \For{$s_1$ to $s_n$}{
            $U(s) \gets \max\limits_{a^\star} \sum\limits_{s^\star \in S} U(s^\star) P(s^\star \vert s,  a^\star)$\\
          }
         
         Return $\pi$ such that for all $s \in \mathbb{S}_I$\\
         $\pi(s) = \argmax\limits_{a^\star} \sum\limits_{s^\star \in S} U( s^\star) P( s^\star \vert  s,  a^\star)$
         
         \caption{Utility Optimization with Ordered MDP's}
\end{algorithm}

\begin{figure}[H]
        \makebox[\textwidth][c]{\includegraphics[width = 4.5in,height=3in]{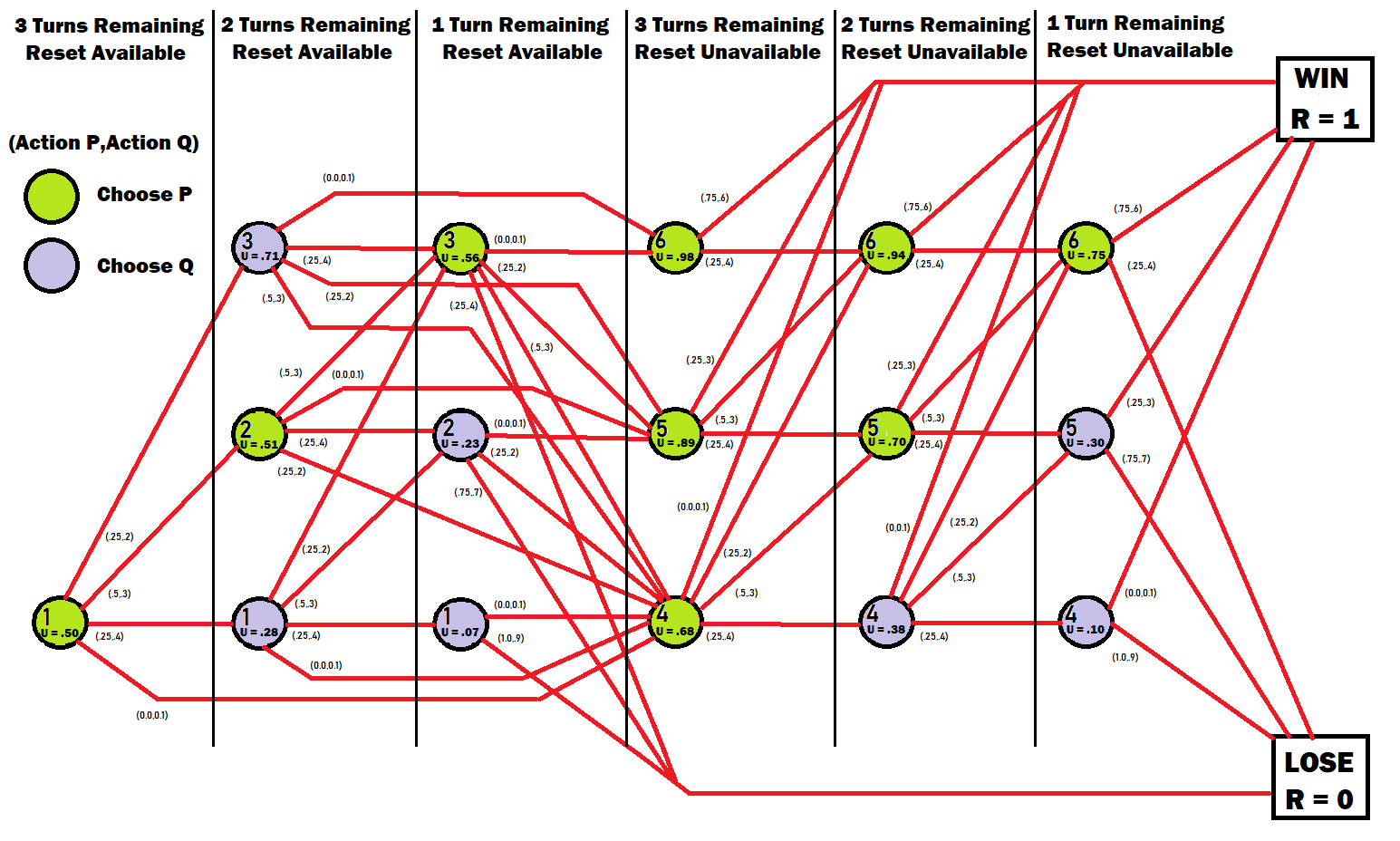}}
        \caption{Toy example with optimal actions labeled along with corresponding utility values.}
        \label{fig4}
\end{figure}

The key to this algorithm is the ordering of the calculations such that utilities are calculated using only terminal states and states where utilities have previously been calculated. In our example, we can see that the intermediate state $s = \{Position = 6,Turns Remaining = 1, Reset Available = False\}$ is dependent only upon the terminal utilities. Thus, the summation used to calculate the action values for this state can be simplified to just the weighted sum of the terminal utilities. Moving one step out, the state $s = \{Position = 6,Turns Remaining = 2, Reset Available = False\}$ relies only on the terminal utilities and the previously discussed state. Thus if we first calculate $U(s = \{Position = 6,Turns Remaining = 1, Reset Available = False\})$, the calculation of $U(s = \{Position = 6,Turns Remaining = 2, Reset Available = False\})$ is again trivial. We can continue this process until reaching the states furthest removed from the terminal utilities, doing one weighted sum for each state/action pair. Thus optimization is achieved without requiring iteration to convergence. Fig.~\ref{fig4} shows this algorithm applied to our example, with the corresponding optimal actions and utilities displayed.

\subsection{Football as a Markov Decision Process}
From here, we develop the previous examples to better model the game of football 
with the aim to answer the question of how to make a decision for the  play  in each state. 
The task then comes down to two parts,
\begin{itemize}
    \item Finding probability distributions that can describe the outcome set of states for any particular state condition and action choice.
    \item Constructing an algorithm which can assign the utilities of each football state, with relevantly assigned terminal utilities.
\end{itemize}
The former will be discussed in Appendix B. The assignment of utilities will be covered in depth in the next section.

\begin{figure}[H]
    \makebox[\textwidth][c]{\includegraphics[width = 4.5in,height=3in]{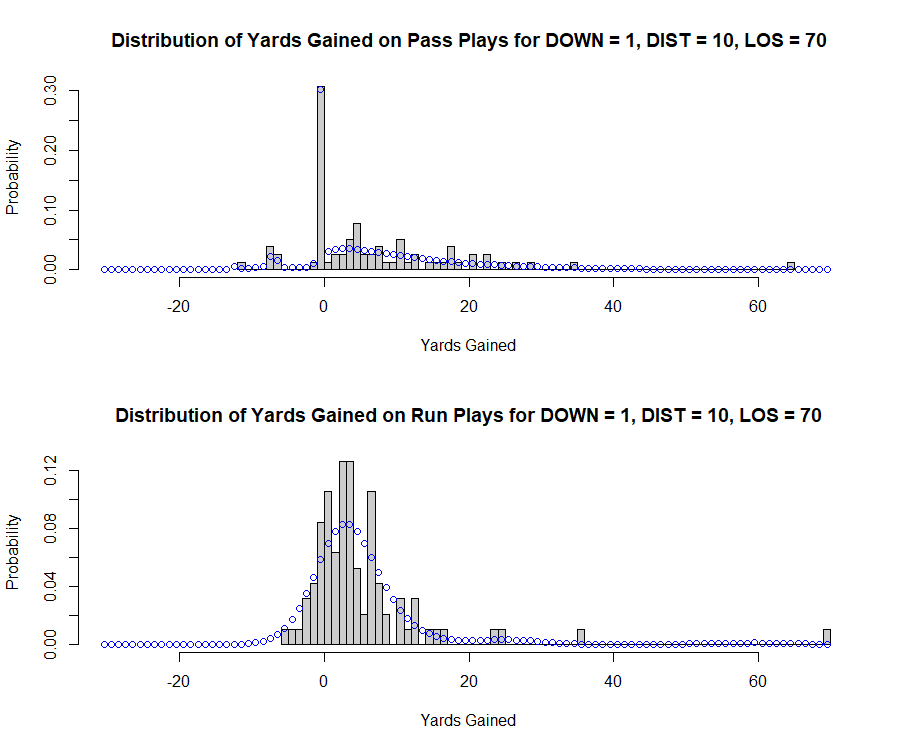}}
    \centering
    \caption{Probability distributions for yards gained on pass and run plays for a 1st and 10 situation from the offensive team's 30 yard line. The grey bars indicate previously observed plays from this scenario, and the blue dots indicate the modeled probability distribution for yards gained.}
\label{fig5}
\end{figure}

Some illustrations of probability distributions of outcomes of plays from real data are presented in Fig.~\ref{fig5}. 

\section{Determination of Utilities}
For this system to be evaluated with an Markov Decision Process (MDP), we must first make the assumptions about the game to be established. The utility calculations will be split between two methods, one for normal game scenarios and one for late game scenarios. The assumptions section will apply to both of these, although assumptions will be written in terms of the normal game scenarios and will be ``tweaked'' when necessary for the late game utility calculations. 

\subsection{Assumptions}
First, we will assume that it is only reasonable to use information available to the coach at the end of the previous play to determine the best decision. Thus, the only information we will use is the following;
\begin{itemize}
    \item Current down ($DOWN$)
    \item Current distance to the first down ($DIST$)
    \item Current line of scrimmage, measured in number of yards away from scoring an offensive touchdown ($LOS$).
\end{itemize}
These will be referred to as the \textit{base preplay information}. Later, we will include both time remaining and score differential into the set of preplay information when we examine late game scenarios.
This assumption excludes us from using information regarding individual players, such as personnel packages being used by the defense in response to what the offense puts on the field. Additionally, any information related to the defense's performance, strategy, or players, along with information relating to the offense not previously listed, including previous plays, will not be used and are considered to be irrelevant in determining a play's success or failure. While it would be helpful to use information relating to previous results within the game, tendencies of the offense/defense, or individual player abilities, the amount of data needed to draw meaningful conclusions would be exponentially larger than what is currently being used. Requiring this much data would hinder a decision maker's ability to make expedient play calls and make it more difficult to perform analysis for a variety of play situations, and therefore we find this assumption to be both helpful and necessary.

Second, we will assume that possessions are independent of one another, conditional on the starting line of scrimmage, score differential, and time remaining in the game. This implies that an offensive team cannot affect the success levels of a future possession, and a defense can only affect the result one possession at a time. A team could affect a future possession using the technique usually referred to as ``flipping the field", which usually is done through a team opting to punt the ball back to the defensive team, pinning them deep in their territory and increasing the probability of ending the drive in a punt. A team could also affect future possessions by scoring (or allowing the other team to score), or by drastically changing the time remaining in the game. For these reasons, we explicitly mention the conditions of line of scrimmage, score, and time to avoid ignoring these important cases.

In reality, there are additional cases where this assumption does not hold. A team could be severely stunted in a future possession from their actions in their current possession in the event of an injury, which would be the main exception to this rule. Another exception would be the idea of setting up plays in the future, such as running a play in a certain style in an early possession, then running a similar play later in the game but tweaking it based off how the defense responded to the play earlier in the game. The assumption could also challenged through the idea of ``establishing the run", a cliche in the game that asserts that running the ball early in the game can improve an offense's effectiveness later in the game, either via future running plays or through play-action passes. The effectiveness of these techniques have been examined by others (Clement 2018, Somers 2020, Baldwin 2018) and certainly may be valid, however we assume that the law of large numbers will dominate over any small difference in individual play success probabilities and will assume the differences that these strategies create are minimal in relation to the overall effects that can be observed using the base preplay information.

Third, we will assume that all sub-plays that make up the whole of ``run" plays have the same distribution for yards gained, and likewise for pass plays. By this, we mean that the probability distribution of yards gained on a I-formation outside pitch is not significantly different than that of a shotgun hand-off run up the middle, which is not significantly different from a quarterback sneak, etc. Similarly, the probability distribution of yards gained on a four verticals (hail mary) pass play is the same as a wide receiver bubble screen. While we are aware there are clear and obvious differences in these play types and therefore know this assumption to be false, the data are not labeled in a way that would permit modeling of individual sub-plays. If data were available with labels for individual sub-play types within run/pass contexts, the methods in this paper could easily be modified to use the new information to create a more robust method of play calling.

Fourth, we will assume that all offenses generate the same probability distributions for each play, and all defenses are equally effective (or ineffective) at stopping each play. By this, we mean that the strongest offense in the league is not significantly different than the worst in terms of the distribution of yards gained for a particular play type, and similarly for defenses. This is obviously also a false assumption, as stronger offenses obviously would generate more yards on average than weaker ones. Like the previous assumption, this was made mainly due to a restriction in the data and could be dropped if the data allowed for a strong modeling method for the probability distributions of play types for an individual team. In many cases, this is actually reasonable. However, in the less common scenarios (long yardage to first down, first down situations that don't have 10 yards to go, fourth down scenarios on the offense's side of the field, etc.) it is necessary that we have accurate estimates of their probability distributions in order to assign utilities properly, therefore we must make assumptions to help fill out the data. Thus, we use the data in aggregate, thereby approximating an NFL average offense in regards to performance. There are previous studies, such as Cafarelli et al. (2012), that suggest that teams do not differ significantly in their offensive abilities on seemingly important metrics such as third-down conversion rate, implying that this assumption should not be considered too outlandish. 

Finally, we will assume that outside of late game (or half) situations, the goal of each offensive team is maximize their expected points over the course of the possession, and the goal of each defensive team is to minimize the same value. In this, we assume that the main goal of teams is to win the game, which is done through scoring (or hindering the other team's ability to score). A case where this assumption might be practically inaccurate might be in long third down scenarios, where one play type might be more likely to score a touchdown or pickup a first down, but the probability of succeeding might be so low that they opt to gain a minimal amount of yardage to push the opposing team's starting field position back on their next drive. Outside of these rare situations, we believe this assumption is valid outside of the end of halves, where a team may have more concern about how much time remains when they finish their drive. This assumption will be dropped when we examine late game decision making, which will be discussed later.

\subsection{Markov Decision Process}
With our assumptions in place, we can now describe how the game of football can be viewed as a Markov decision process. For a review off MDP, see, for example, White and White (1989). Here we will use the MDP to find the utilities that should be assigned to each state, along with the reward that can be gained by choosing an action at each state.

We will assume that the decision making policy for the process is known; always choose the action which maximizes the expected utility of the future state options. While there may be some practical benefit to be gained from using an element of deception in decision making, potentially choosing sub--optimal plays to influence the defense to respond in a way that improves the offense's probability of succeeding on future plays, we are operating under the assumption that the probability distribution of a particular play is static, and therefore an offense cannot change the probability distributions of play calls later in the game. For practical purposes, one could assume that if a team has a ``smart'' quarterback, the player could quickly recognize when the opposing defense is prepared to stop the called play. While the data here will help suggest a play that should be considered the best play on average, via the law of large numbers, we do not argue that these decisions are infallible and leave open the opportunity for shrewdness when it is deemed necessary by an informed decision maker. That is, for a given play, there is overwhelming evidence on the field the current action should be chosen at the moment.

A MDP has four components;
\begin{itemize}
    \item The set of states, $\mathbb{S}$, referred to as the State Space;
    \item The set of actions, $A$, referred to as the Action Space;
    \item The probability that action $a$ in state $s$ at play $k$ will lead to state $s^\star$ at play $k + 1$, $P(\mathbb{S}' = s^\star \mid \mathbb{S} = s, a_k = a)$, referred to as the Transition Probability;
    \item The immediate reward after transitioning to state $s$, $R(s)$, referred to as the Immediate Reward.
\end{itemize}
As we have already discussed, the state space in football can be characterized by base preplay information, $s = \{DOWN,DIST,LOS\}$, and the terminal states of an offensive touchdown, offensive field goal, defensive safety, and defensive touchdown. The action space is fully characterized by four elements, $A = \{RUN,PASS,FG,PUNT\}$, where only the former two are reasonable options in most states. The transition probability is carefully determined using data, which is described in Appendix B. 

Regarding the Immediate Reward, we must split the discussion between terminal and non--terminal states. For terminal states, the Immediate reward can have a clear interpretation; the amount of points gained (or lost) by the offense for entering that state. The terminal states that are available for any particular offensive play, along with points gained (or lost, indicated by a negative) are as follows;
\begin{itemize}
    \item $R(TD) = 7$
    \item $R(FG) = 3$
    \item $R(SAFETY) = -2$
    \item $R(DefTD) = -7$
\end{itemize}
It should be noted that a touchdown play actually results in only 6 points automatically, and the offense can either make it 7 with an extra point kick or could choose to attempt a 2-point conversion to add 0 or 2 points on a failure or success, respectively. However, the data shows that the expected points for an actual touchdown tends to be around $6.95$ (Morris 2017), given a two-point conversion tends to succeed about 45\% of the time and extra points are made around 90\% since they were pushed back to the $15$ yard line. Thus, we will approximate this with $7$ and simplify computations. This will be handled robustly when we examine late game scenarios.

We choose here to assign the immediate reward value as the points gained in these states as they allow for a clear utility value interpretation. While other values could be used, different values would both lack the direct interpretation that using actual points provides and could potentially damage the connection to the game and therefore may be more of a reflection of personal beliefs of the game rather than an objective utility calculation. For example, a coach may say that a defensive touchdown is more damaging to an offense than the positive gain an offense gets from an offensive touchdown, therefore the utility of a defensive touchdown should actually be more negative than $-7$, or conversely the offensive touchdown should be worth less than $+7$. While this may or may not be accurate in terms of how a team actually responds to positive and negative game outcomes, for a model to be mathematically sound, it must have a clear connection to the both the data it uses and the game it represents. Therefore, utility values should be chosen in a manner that limits subjectivity, and thus we opt to use values that can be observed within the context of the game.

Having assigned the immediate reward values for the terminal states, we must now examine how to determine the immediate rewards for non-terminal states. Considering the goal is to score a touchdown on every offensive drive, the utility of a particular state must be the expected points one would get if following an optimal set of actions for the rest of the drive. Therefore, depending on the base preplay information, the utility of a state can be directly calculated using a probabilistically determined expected points model, where the expected points are recursively calculated backwards through the states.

Thus, with this context in mind, the task of understanding the immediate reward becomes clear. Assuming that there is sufficient time remaining in the half, there is no immediate reward to be gained on any particular play that is not incorporated in the calculation of the utility from future states. To say that any state has a non-zero immediate reward would imply there is an inherent value to being at a certain $DOWN$, $DIST$, and $LOS$ that is independent of the expected value of scoring from that position, and therefore if the offense did not end up scoring on the drive, they would still walk away from the drive with some ``reward" despite the score not changing in their favor. From an analytical perspective, while there may be emotional benefits for having a ``successful" drive resulting in no points, it would be improper to attempt to estimate intangible gains with no data to consistently substantiate any particular effects. However, we specifically mention the exclusion of late game/half scenarios from this discussion, as it is easy to argue that there would be value in running the clock. This will be discussed in detail in the next section.

Thus, we will set $R(s) = 0$ for all intermediate states $s \in \mathbb{S}_I$. When we examine the late game scenarios, we will continue to set this value to zero, opting to change our state space definition to incorporate time and score differential to simplify the computational process and keep our assumptions consistent across methods. Note that in no part of this process are we including a discount factor $\delta$, implying there is no benefit for scoring at a faster or slower pace. This holds for the late game scenarios as well, as we will view the utilities in terms of winning an losing at the end of the game, and therefore there is no reduction in value for reaching the $WIN$ state at a slower pace.

Finally, it is worth discussing the Transition Probability in terms of which states can be transitioned to with nonzero probability. For any particular state $s$, there are a maximum of 202 possible next states $s^\star$. This applies for all $DOWN$, $DIST$, and $LOS$ combinations in the game, ignoring SD and TIME. The potential next states are as follows;
\begin{itemize}
    \item Offensive Touchdown
    \item Offensive Field Goal
    \item Defensive Touchdown
    \item Defensive Safety
    \item Offense retains possession of the football without a scoring event, and has the ball at $LOS = 1,\dots,99$
    \item Defense obtains possession of the football but does not score, and gets the ball at $LOS = 1,\dots,99$
\end{itemize}
For action choices of run and pass, only 201 of these states are obtainable, as a field goal would be removed. On a field goal attempt for the offense, we will view the future states as either ``Field Goal Made" or ``Field Goal Missed", the latter of which can actually be viewed as the defense obtaining possession of the football at the current $LOS$ (or $100-LOS$ in the defense's context) with $DOWN = 1$ and $DIST = 10$. For punts, the set of future states removes all options where the offense scores or retains possession of the ball, leaving the defensive touchdown, safety, and defensive possession future states to calculate. We included a small point mass of the offense retaining possession of the football in the case of a muffed punt, giving the offense a first down at the yard line of the recovered fumble. We truncated all of the potential recovery yard lines to that of the mode of the punt distance distribution, assuming that while the fumble could be recovered anywhere on the field, it most often occurs at the location of the punt's landing location, which can be approximated by the peak of the punt return distribution.

In terms of the plays where a scoring event does not happen, it is straightforward to see how the states advance. Given that the down on play $k$, denoted $DOWN_k$, is less than 4, when the offense runs a play where they keep possession and do not score (most plays), they will gain a certain number of yards which we will refer to as $GAIN_k$. In the case that $GAIN_k$ is less than $DIST_k$ (they gained less yards than they needed to obtain a first down), the future state becomes $DOWN_{k+1} = DOWN_k + 1$, $DIST_{k+1} = DIST_k - GAIN_k$, and $LOS_{k+1} = LOS_k - GAIN_k$. Note that $GAIN_k$ must not be positive in this scenario. In the case that $GAIN_k$ is greater than $DIST_k$ (they gained enough yards for a first down), the future state becomes $DOWN_{k+1} = 1$, $DIST_{k+1} = min(10,LOS_{k+1})$, $LOS_{k+1} = LOS_k - GAIN_k$. Notice here that the new distance either becomes 10 or the new distance to the end zone, thus becoming first and goal if the offense moves to within the 10 yard line. Remember that if $GAIN_k = LOS_k$, then the offense gained exactly enough yards to score a touchdown and this would be a scoring play. In the case that $DOWN_k = 4$, the setup of future states would be the same for when $GAIN_k$ is greater than $DIST_k$, but for cases where it is less, the play would result in a turnover on downs. In this scenario, the defense obtains control of the ball. Therefore, are no results where the offense retains control of the ball at a yard line where $LOS_{k+1} = LOS_k - GAIN_k$, thus the future state options are actually truncated on fourth down. This covers all of the 99 different options for the offense controlling the ball without scoring.

In this context, we can view the defense obtaining possession of the football states as semi-terminal states, denoted $\mathbb{S}_M$. We use the semi-terminal nomenclature to indicate that it does indicate the end of the offense's possession, however it does not necessarily mean that the calculation of utilities is finished, as it is important to know the expected utility of each future state to determine that of the current state. With the scoring event terminal states, there is a clear value associated that can be used for utility calculation. But for these semi-terminal states, the defense has taken control of the ball via a turnover (interception, fumble, or turnover on downs) or punt and become the new offense and now has the opportunity to take the ball and score themselves. Thus they would have their own set of future states that need to be calculated. This would then bring into effect the possibility of the defense turning the ball back over to the offense, bringing into account a new set of semi-terminal states. This could continue ad infinitum, and therefore should be truncated to allow for convergence. To deal with this recursive loop, we must first examine what these future states will be. In every case, the 99 semi-terminal states that give the defense possession of the football are the following:
\begin{itemize}
    \item $DOWN = 1, DIST = 10, LOS = 10,\dots,99$, or
    \item $DOWN = 1, LOS = 1,\dots,9, DIST = LOS$.
\end{itemize}
Thus, if the defense gets possession anywhere other than within 10 yards of scoring a touchdown, they will have 1st and 10. If they do obtain the ball within 10 yards of scoring, they have first and goal. Thus, the future semi-terminal states are consistent for all states. Additionally, since we assumed that all offenses are equivalent and all possessions are independent, we can start our calculations by assuming some reasonable fixed set of utility values for each of these semi-terminal states, and update them periodically. Thus, when the 1st down states are called from an offensive perspective, we will calculate them as normal. However, when they are called as semi-terminal defensive states, we will use their current values, multiplied by negative one to imply a negative result for the offense. This allows the system to reach convergence and the utility values to be obtained. Note that while the probability of ending in each of these states may change for different action choices, the resulting state options are equal for run and pass plays.

Thus, we now have the information needed to start calculating utilities. At any particular state, the value of a particular action $a$ at a given state $s_k$, $V(s_k,a)$, along with the utility of a particular state, $U(s_k)$ can be defined as follows:
\begin{align*}
    V(s_k,a) &= \sum\limits_{s_{k+1} \in S} U(s_{k+1}) P( s_{k+1} \mid s_k,  a)\\
    U(s_k) &= \max\limits_{a^{\star}} V(s_k,a^{\star})
\end{align*}
with
$    U(s = \text{Offensive Touchdown}) = 7$, $U(s = \text{Offensive Field Goal}) = 3$,  $U(s = \text{Defensive Safety}) = -2$ and
    $U(s = \text{Defensive Touchdown}) = -7$.
For consistency in terminology, we will use the term utility when referring to states, and value when referring to actions.

Using this setup and knowledge of the game of football, we know that every state will have its utility determined by the maximum value of the actions that can be taken at that state, and therefore determining the best play at each state requires simply choosing the action that maximizes the state utility. The action values are then determined by the probability weighted average of the future state utilities, with three of the future states always being terminal along with 99 semi-terminal states of the defense obtaining possession. Thus, at each state, there are a maximum of 99 future state utilities that need to be calculated before the original utility can be evaluated, which will eventually all cascade down to terminal states as the offense approaches the end zone.

The following algorthm was used to complete this process of calculating the utilities. The algorithm incorporates elements from both of the algorithms discussed in the simpler examples, and achieves convergence very quickly (less than 10 iterations).

\begin{algorithm}[H]
        \SetAlgoLined
        \KwResult{A deterministic policy, $\pi$, that designates the optimal action at each intermediate state }
         Choose a small threshold $\theta > 0$, determining accuracy of estimation\;
         Initialize $U(s)$ for all $s \in \mathbb{S}_I$ and $\mathbb{S}_M$ arbitrarily\;
         Initialize $U(s)$ for all $s \in \mathbb{S}_T$ as designated terminal rewards\;
         Initialize $\Delta > \theta$, arbitrarily;\\
         Order $\mathbb{S}_I$ such that $s_j \in \mathbb{S}_I$ depends only on $s_1$, \dots, $s_{j-1}$, $\mathbb{S}_M$, and $\mathbb{S}_T$\\
        \While{$\Delta > \theta$}{
          \For{$s_1$ to $s_n$}{
            $u \gets U(s)$\\
            $U(s) \gets \max\limits_{a^\star} \sum\limits_{s^\star \in S} U( s^\star) P( s^\star \vert  s,  a^\star)$\\
            $\Delta \gets \Delta + (U(s) - u) ^ 2$\\
          }
         \For{each $s$ in $\mathbb{S}_M$}{
            $u \gets U(s)$\\
            $U(s) \gets \max\limits_{a^\star} \frac{\sum\limits_{s^\star \in \mathbb{S}: s^\star \neq s } U(s^\star) P( s^\star \vert  s,  a^\star)}{1 - P( s \vert  s,  a^\star)}$\\
            $\Delta \gets \Delta + (U(s) - u) ^ 2$\\
          }
          }
         
         Return $\pi$ such that for all $s \in \mathbb{S}_I$\\
         $\pi(s) = \argmax\limits_{a^\star} \sum\limits_{s^\star \in S} U(s^\star) P( s^\star \vert  s,  a^\star)$
         
         \caption{Value Iteration with Semi-Ordered MDP's}
\end{algorithm}

\subsection{Late Game Utility Evaluation}
While the methods described in the previous section operate on assumptions that are reasonable in most in-game situations, team strategies become much different near the end of the game. A team with the lead reasonably becomes less concerned with scoring more points and more concerned with conserving their lead and draining the clock. A team at a deficit alternatively will still be concerned with scoring but will consider how their actions affect the clock. In both cases, the utilities gained for scoring events might not align with the actual points gained for the scoring event. For example, while a team would always be very satisfied with scoring a touchdown, a team down 2 points with 30 seconds remaining will devote their efforts into getting into field goal position. This implies that the team does not care particularly about the margin of victory, but would be satisfied with a field goal to win the game, implying that the utility of a field goal is much closer to that of an offensive touchdown. Thus, we must alter our methodology to fit this new mindset.

To fit this context, we will need to use extra information to make our decisions, therefore we will add the following pieces to our base preplay information:
\begin{itemize}
    \item Current score differential, measured as offensive team's score minus defensive team's score (SD); and
    \item Current number of seconds remaining in the game (TIME).
\end{itemize}

These two new information pieces along with our base preplay information will be referred to jointly as the \textit{adjusted preplay information}. The last piece information, TIME, is somewhat questionable in its designation as ``preplay" information, in that the clock may or not be running (actively decreasing the value of TIME) when the coach makes the play call. However, it is safe to assume that the coach is aware of the clock and will be able to estimate at a reasonably accurate level the amount of seconds remaining at the snap of the next play, which is the value we will use to make computations. Using this set of values, it is now possible to adjust our methods to find the utilities necessary for making decisions in late game scenarios.

First, we define ``late game scenarios" as those within five minutes of the end of the game. It would be reasonable to use the utilities obtained through this method and apply them to the end of the first half as well, although they should be done with caution as the implications of being behind at the end of a half are not as severe as being behind at the end of the game, therefore the actions taken to avoid a loss at the end of the game might not be the same as those taken at the end of a half.

Next, we will adjust our state space to now include our full adjusted preplay information. With this new set of states, the utility that is gained in the terminal states is no longer how it was originally defined. In fact, the terminal states themselves are no longer terminal, as we need to view the game as a whole rather than as individual possessions. Therefore, we will redefine the terminal states as those where $TIME = 0$, as this is the true end of the game in this context. Since teams only need to have a lead of one point to win the game, we will assign the terminal utilities as follows;
\begin{align*}
    U(s \mid SD > 0, TIME = 0) &= 1\\
    U(s \mid SD = 0, TIME = 0) &= 0.5\\
    U(s \mid SD < 0, TIME = 0) &= 0.
\end{align*}
Thus, instead of viewing the utilities in terms of scoring points, we have changed to viewing them in terms of winning or losing the game. These terminal values can be seen as what happens when the game ends: if the team has a lead, it gets a maximum utility of 1, if it is losing, then it gets a minimum utility of 0. In the case of the score differential being equal to zero, we simply call the utility equal to 0.5 rather than simulating an overtime event to attempt to settle the tie. There are many studies of overtime that have been conducted (Zauzmer 2014, Leake and Pritchard 2017); this will not be done in this paper for simplicity, and we will assume that with no a priori knowledge of the teams, teams tied at the end of regulation will win roughly 50\% of the time. 

From here, the task becomes to again determine the state utilities and action values which are now both time and point differential dependent. This will require a few modifications to the previous system:
\begin{itemize}
    \item When scoring events occur, the score differential must be modified accordingly and sent to the proper transition state;
    \item Upon selecting an action for each state, the $TIME$ factor must be adjusted probabilistically according to the selected action; and
    \item The cycle of selecting actions and evaluating states will continue until a terminal state is reached, which now is dependent upon $TIME$.
\end{itemize}
Making these modifications requires some delicacy. For the first, instead of scoring events ending the computations (assuming $TIME > 0$), the system will now proceed into a kickoff transition state, awarding possession of the ball to the team that allowed the scoring event (outside of safety events, in which the scoring team gains possession). For the second adjustment, we must now take into account the distribution of time for each action choice when updating the utility values. The details of how the time distributions are calculated are available in Appendix B.9. Regardless, the algorithm will determine the proper time adjustments necessary for each action and influence the decisions to be made for each play accordingly. Finally, the system of computing utilities will continue until the new terminal condition has been reached, and making the final step of the process to be determining whether a string of actions resulted in a team winning or losing the game. With this new system, the state and action utilities were computed, and their results will be discussed in Section 4.2. 

The following algorithm shows how the utility and optimal play call can be determined for a single state using the methods described in this section. Due to the enormity of the state space, we are no longer able to exhaustively iterate through all the states. However, if utility values are initialized reasonably (states with positive score differentials have initial values greater than $\frac{1}{2}$, etc.) and updates are stored intermittently, the accuracy of the found utilities will increase and values will eventually converge.

\begin{algorithm}[H]
        \SetAlgoLined
        \KwResult{An action choice to take at input state $s$, along with updated utility vector $U$}
        \KwIn{Current utility vector $U$ and state to calculate optimal action $s$}
         \For{each $a \in A$}{
              $\mathbb{S}_a \gets s^\star \vert P( s^\star\vert s, a) > 0$
         }
         Initialize $S^C \gets \varnothing$\\
         \For{each $a \in A$}{
             \For{each $s^\star \in \mathbb{S}_a $}{
                \uIf{$s^\star \notin S^C$}{
                    $U(s^\star) \gets \max\limits_{a^\star} \sum\limits_{s^{\dagger} \in S} U( s^{\dagger}) P( s^{\dagger} \vert  s^\star,  a^\star)$\\
                    Append $s^\star$ to $S^C$
                }
             }
             $V(s,a) \gets \sum\limits_{s^\star \in \mathbb{S}_a}  U( s^\star) P( s^\star \vert s,  a)$
         }
         $U(s) \gets \max\limits_a V(s,a)$
         
         Return $a$ such that
         $a = \argmax\limits_a V(a,s)$, and utility vector $U$
         
         \caption{Single State Optimization for MDP's with Dynamic Programming}
\end{algorithm}

\section{Results}
This section details the utility values we computed.
The discussion will be divided between the two types of utility values derived, i.e. those calculated for normal game circumstances and those calculated for late game scenarios. We will first focus on their interpretation in the mindset of a play caller, and then look at them from the perspective of the sports analytics community.

\subsection{Normal Within Game Utilities}
\begin{figure}[H]
    \makebox[\textwidth][c]{\includegraphics[width = 4.5in,height=3.5in]{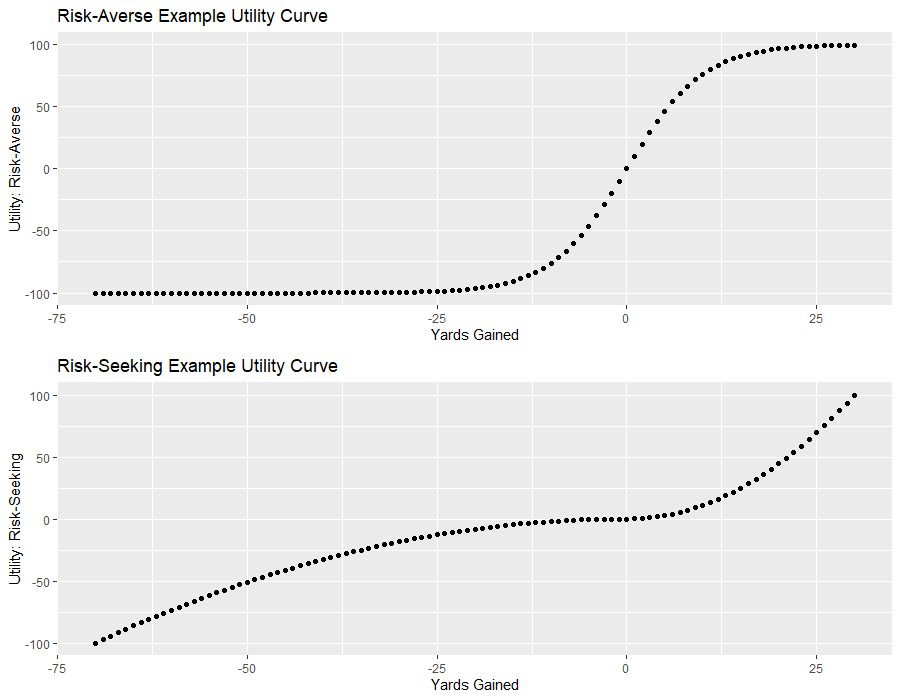}}
    \centering
    \caption{Utility curves for two differing mindsets, with the first representing a risk-averse decision maker, and the second representing a risk-seeking decision maker. One would expect the first decision maker to be more prone to accepting smaller gains with a higher probability of success, while the second to be more likely to be willing to risk failure for the potential of larger gains.}
    \label{UtilMindsets}
\end{figure}

With the utilities calculated, we can now view individual plays and determine the ``optimal" play call in each scenario. This process comes down to simply selecting the action for each state which maximizes the state's utility.
Additionally, decision theory (Hirshleifer and Riley, 1992) gives us insight of the mindset one should have based off the shape of the utility curve. For any state, we can plot the set of utility values with respect to the resulting line of scrimmage, and thus understand how a scenario should be viewed. With this methodology, it would be helpful to observe utility curves for general  plays, in order to get an idea of how the data indicate one should respond for each circumstance. Fig.~\ref{UtilMindsets} shows two differing theoretical utility curves for the same scenario, that could potentially result in different play calls.

It should be noted that this discussion is to provide a general overview of each of the circumstances, and the idea of having an aggressive or passive mindset in any scenario is not necessary when the full data is available, as the optimal play call can be determined and should not be diverted from for the sole reason of matching predetermined aggressive/passive tendencies. Additionally, only some plays will be discussed in this section, as the goal is to point out the capabilities of the method rather than exhaustively discuss play calling.

\subsubsection{1st Down}
\begin{figure}[H]
    \makebox[\textwidth][c]{\includegraphics[width = 4.5in,height=3in]{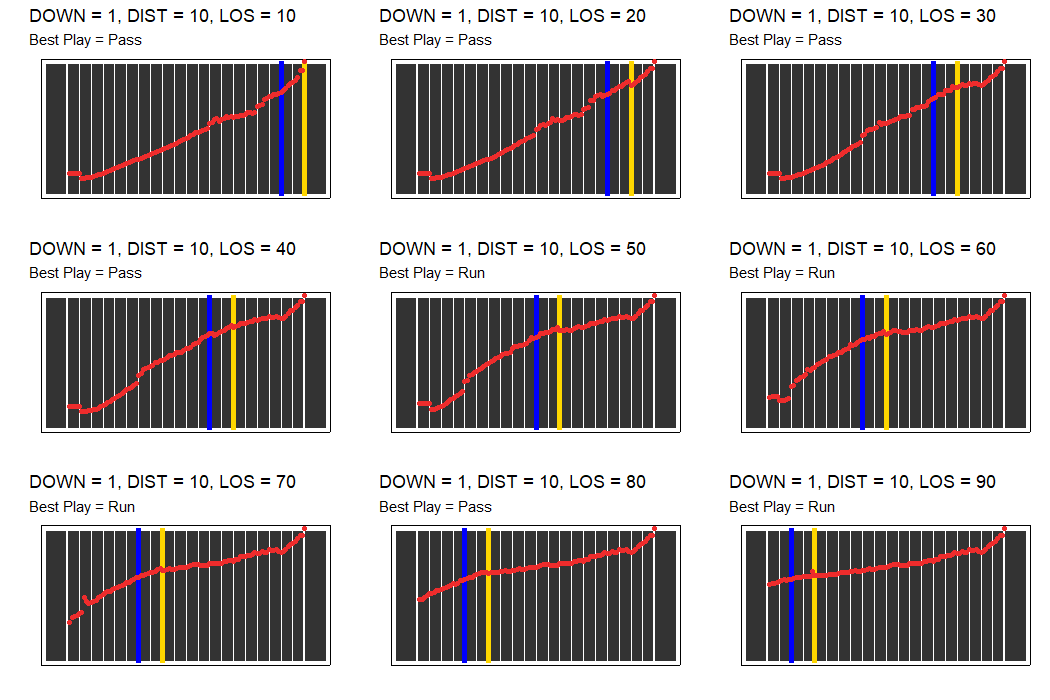}}
    \centering
    \caption{Utility curves for first down plays.}
    \label{fig7}
\end{figure}

Fig.~\ref{fig7} shows the utility curves for a handful of first down plays, spread across the field, along with the optimal play call for each of the plays. For each of these plots, the y-axis scale goes from negative to positive seven. The $x$-axis indicates the line of scrimmage, drawn to mirror the look of a football field, where the further right you go on the plot, the closer the offense is to scoring a touchdown. The blue line indicates the current line of scrimmage, and the yellow line indicates the yard line the offense must reach to gain a first down. 

In these plots, the play recommendations tend to be passes when closer to scoring, and runs when further from the end zone. We tend to see a slight dip in the utilities at the yellow line, indicating that it is generally more favorable to an offense to have 2nd down and short over a new first down. This may be surprising, but it also implies that an offense benefits from having more attempts to gain yardage, and gaining a first down too quickly without a significant gain in yards may lower the team's expected score. The shape of the curves tend to be fairly linear between the current line of scrimmage and the first down line, indicating a relatively risk neutral mindset.

\begin{figure}[H]
    \makebox[\textwidth][c]{\includegraphics[width = 4.5in,height=3in]{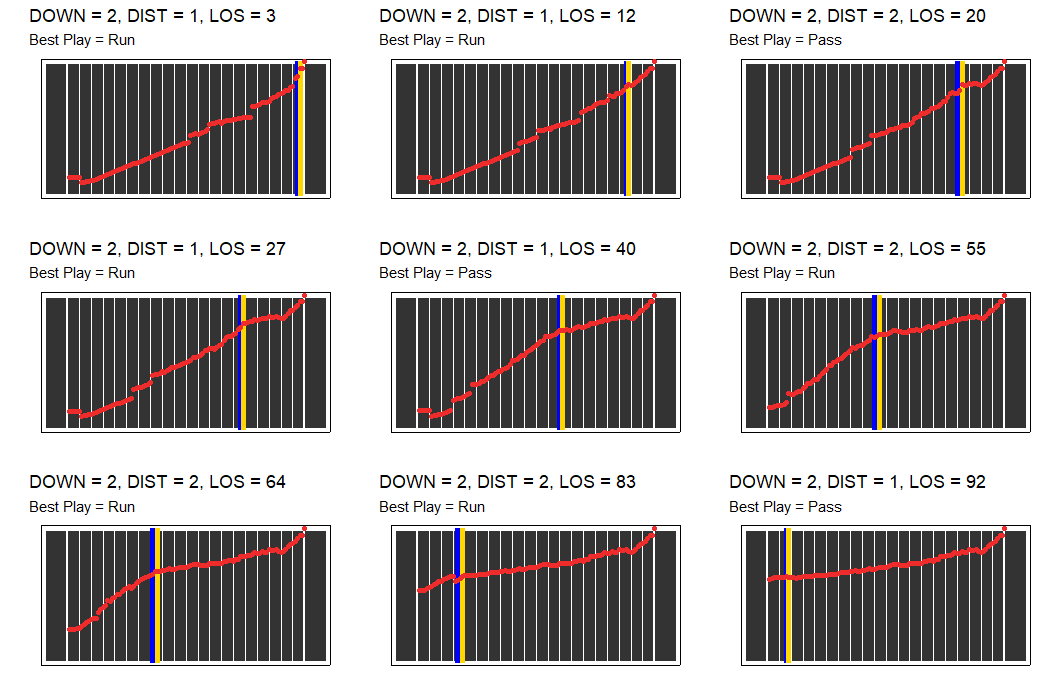}}
    \centering
    \caption{Utility curves for second down and short plays.}
    \label{fig8}
\end{figure}

\subsubsection{2nd and Short}
Fig.~\ref{fig8} shows utility curves for several 2nd and short scenarios, defined as 2nd down and 2 yards or less, the exact yardage of which is explicitly labeled on each plot.

The play recommendation tends to be run, indicating that it is often better for a team to pick up the first down (and potentially additional yards) sooner rather than risk an incompletion on a pass play. Additionally, we notice that the parts of the graph to the left of the first down line tend to trail off quicker towards negative utility values than those in the first down utility plots. This is to be expected, as not getting to the first down line on first down is much less of an issue than not achieving it on second down, as you have more opportunities to get there later. The shapes of the curves again tend to be fairly linear, continuing to indicate a risk neutral mindset.

\subsubsection{2nd and Long}
\begin{figure}[H]
    \makebox[\textwidth][c]{\includegraphics[width = 4.5in,height=3in]{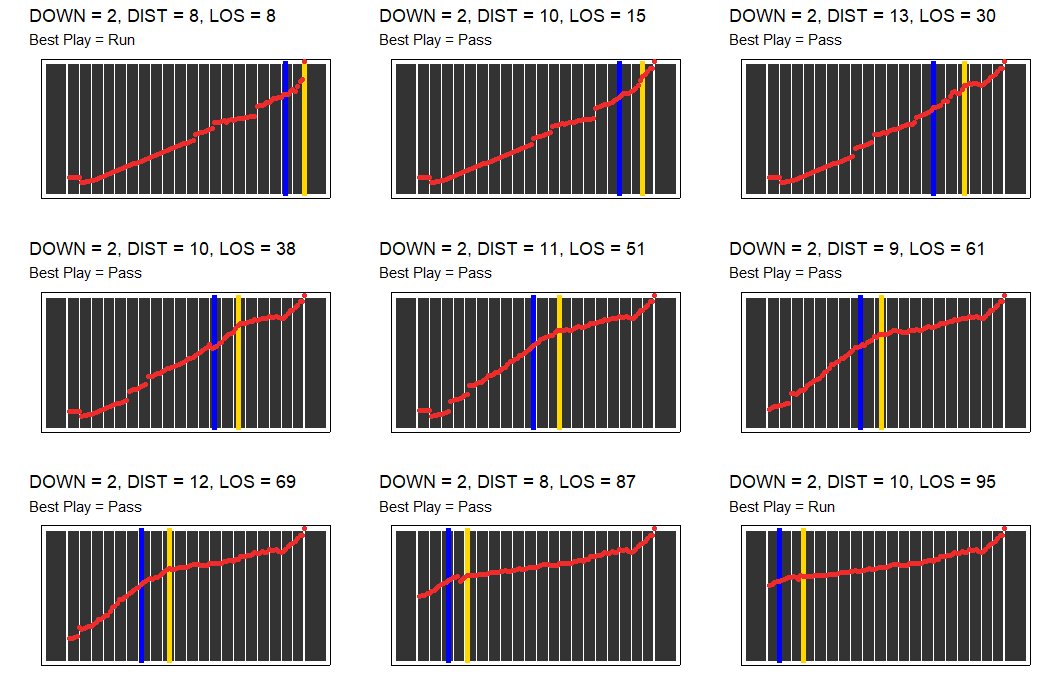}}
    \centering
    \caption{Utility curves for second down and long plays.}
    \label{fig9}
\end{figure}

Fig.~\ref{fig9} shows utility curves for several 2nd and long scenarios, defined as 2nd down and 8 yards or more, the exact yardage of which is explicitly labeled on each plot. 

Outside of the two scenarios within 10 yards of either end zone, the play recommendation is to pass. This is complemented by the shape of the curves, which tend to be fairly concave up between the line of scrimmage and line to gain, indicating a risk-seeking mindset. In these scenarios, the offense is encouraged to increase their probability of a no-gain or turnover in order to gain a first down, and the pass play aligns with this mindset.

\subsubsection{3rd and Long}
\begin{figure}[H]
    \makebox[\textwidth][c]{\includegraphics[width = 4.5in,height=3in]{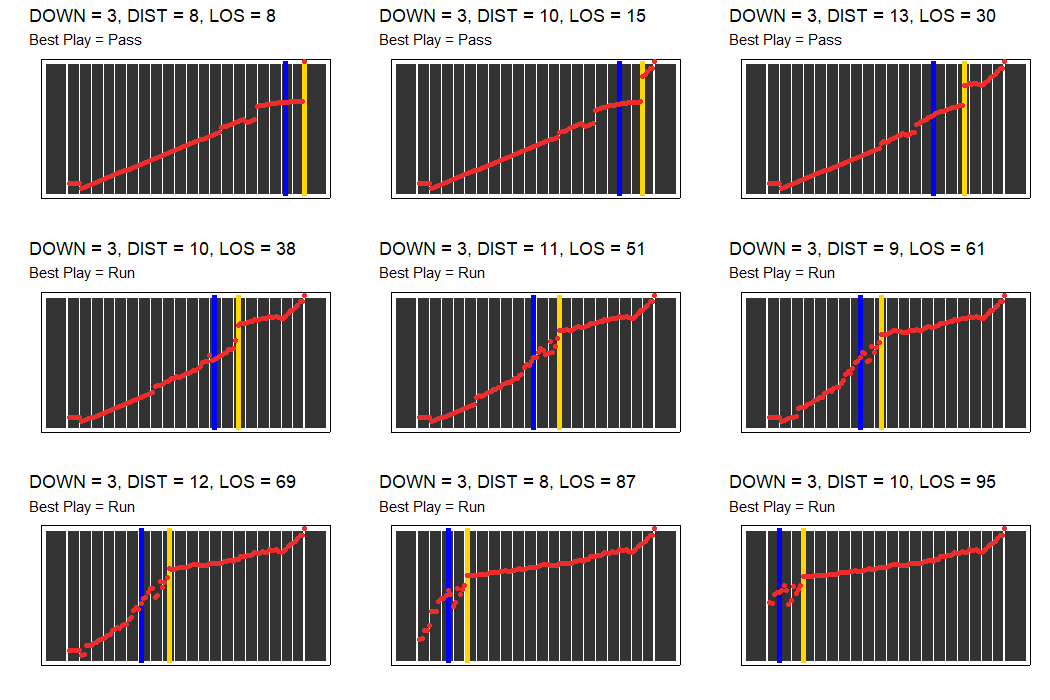}}
    \centering
    \caption{Utility curves for third down and long plays.}
    \label{fig10}
\end{figure}

Fig.~\ref{fig10} shows utility curves for several 3rd and long scenarios. 

Interestingly enough, the play recommendation is only to pass in the cases where the offense is relatively close to scoring a touchdown. Typically considered a passing down, many of these third and long situations recommend running the ball for a variety of reasons. One of the reasons is that often the utilities indicate that the team plans to go for it on fourth down if the team can get into a 4th and short situation, and therefore running the ball may maximize this opportunity. Another reason is that the team may plan to kick a field goal or punt the ball, and running in these scenarios increase the probability of making the field goal or pinning the opponent deep.

One of the biggest factors that may not have been expected is that run plays often have more success than passing plays on third and long scenarios. Due to the historical tendencies of teams to pass the ball on third down, defenses have adjusted to plan to stop the pass, believing that even if the offense chooses to run, the extra defensive backs in the game should be able to break off from their pass-coverage and tackle the runner short of the line to gain. While often they may be correct, this tends to allow for run plays to go for more yards on average in third and long scenarios, making it a relatively advantageous play for an offense.

Finally, in each of these graphs, we tend to observe a very large gap in the utility values from coming one yard short of a first down to getting to the line to gain. This matches with our understanding of the game, as gaining 7 yards on 3rd and 8 will often lead to the offense punting or kicking a field goal, while gaining 8 yards would result in a first down, and therefore give the opportunities several more opportunities to attempt to score a touchdown. This indicates that our utilities line up with our understanding of the game fairly well and can be trusted.

\subsection{Late Game Utilities}
Similar to the normal utility values, once calculated, we can determine best plays as well as tendencies based off the game scenarios. Rather than focusing on the $DOWN, DIST,$ and $LOS$ metrics and how they relate to the utilities, here we will focus on how the score differential relates to the derived values. To do this, we will look at what play is recommended in an array of scenarios taken with a minute or less remaining, picked randomly from the set of possible states (stratified by $DOWN$ and limited to reasonable values of $DIST$). These scenarios were held constant through the following subsections, with the only variable changing being the score differential. Note that these values were not trained to convergence, and therefore the utilities are different from their true values and recommendations may change in future iterations. However, we do expect trends that would be displayed in a real game situations to hold for the majority of the scenarios.

\subsubsection{Large Lead}
The following table shows the play call distribution for decisions made in late game scenarios with a large lead, defined as having a lead of greater than 8 points, which would require an opponent at least two scoring plays to take the lead.

\begin{figure}[H]
    \makebox[\textwidth][c]{\includegraphics[width = 4.5in,height=3in]{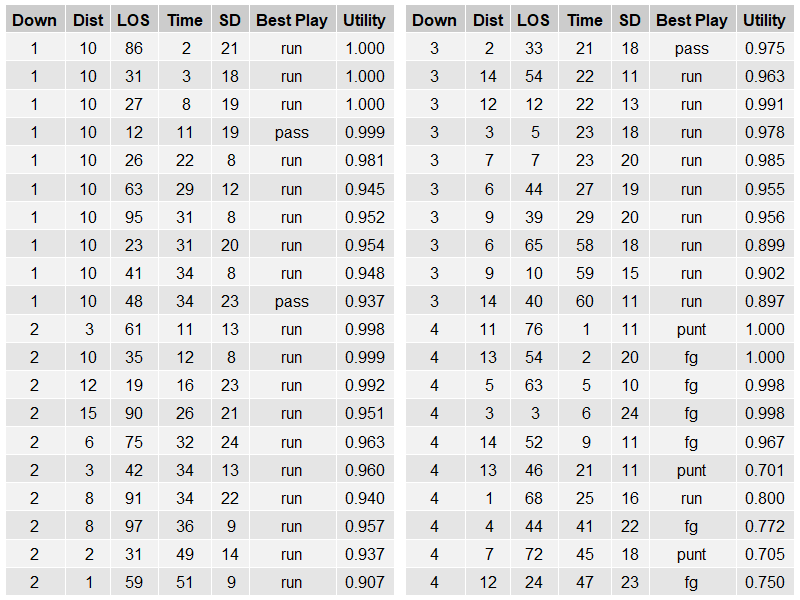}}
    \centering
    \caption{Table of optimal play calls and corresponding utilities for scenarios where the offensive team has a large lead.}
\end{figure}

As expected, most of the play recommendations are to run the ball, implying the best option is to drain the clock. We see exceptions to this on fourth down, where kicking the ball away or attempting a field goal become more reasonable options. Occasionally, we will see different play recommendations. However, closer examination of these plays shows that in most cases, the probability of winning is so high, that no matter the decision, the game will end in a victory.

\subsubsection{Large Deficit}
The following table shows the play call distribution for decisions made in late game scenarios with a large deficit, defined as trailing by a score differential of greater than 8 points.

\begin{figure}[H]
    \makebox[\textwidth][c]{\includegraphics[width = 4.5in,height=3in]{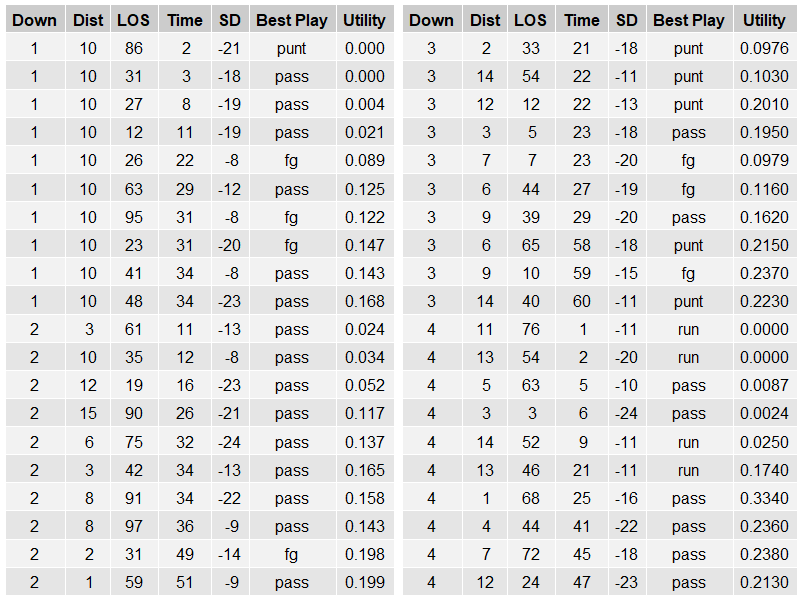}}
    \centering
    \caption{Table of optimal play calls and corresponding utilities for scenarios where the offensive team has a large deficit.}
\end{figure}

Conversely to the large leads, we see pass plays are the majority of recommendations. This follows what we observe in actual games, as passing will limit the draining of the clock and open up more opportunities to score quicker. Again, we will occasionally see kicking type plays recommended, implying that no matter what action we take, the game is likely over.

\subsubsection{Small Lead}
The following table shows the play call distribution for decisions made in late game scenarios with a small lead, defined as having a lead of between 1 and 3 points, which would require an opponent to score at least a field goal to match or take the lead.

\begin{figure}[H]
    \makebox[\textwidth][c]{\includegraphics[width = 4.5in,height=3in]{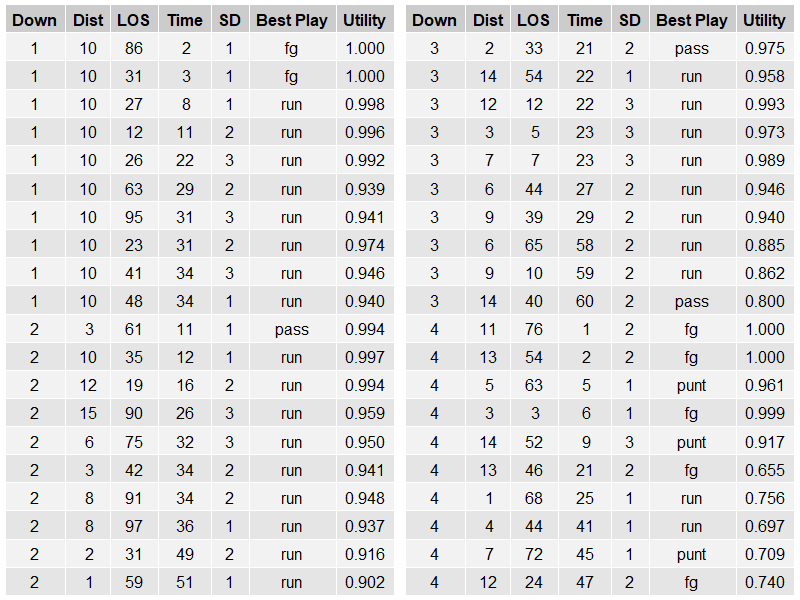}}
    \centering
    \caption{Table of optimal play calls and corresponding utilities for scenarios where the offensive team has a small lead.}
\end{figure}

While the recommendations get a little more diversified, we still tend to see run plays recommended, particularly on earlier downs when the main goal is draining the clock rather than getting another first down. We see that the utilities for the scenarios have dropped from their values in the large leads, indicating that victory is less secured. 

\subsubsection{Small Deficit}
The following table shows the play call distribution for decisions made in late game scenarios with a moderate deficit, defined as trailing by a score differential between 1 and 3 points.

\begin{figure}[H]
    \makebox[\textwidth][c]{\includegraphics[width = 4.5in,height=3in]{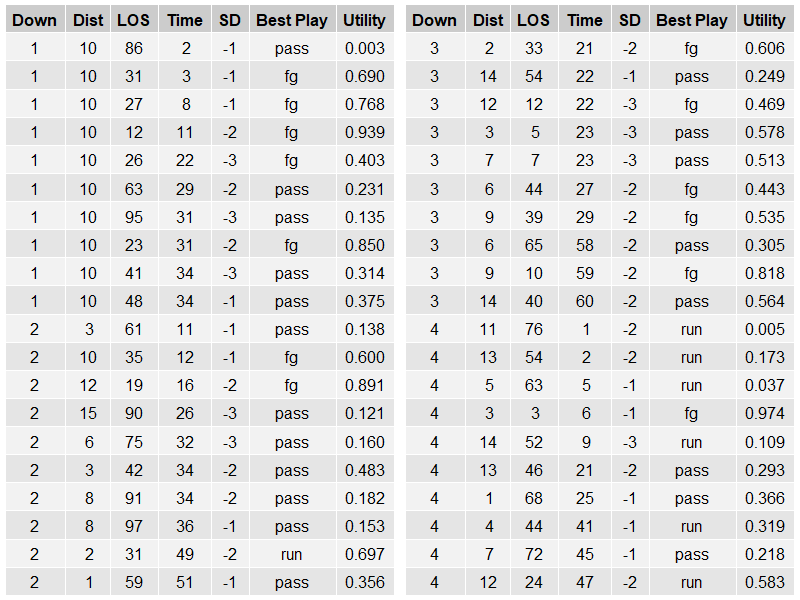}}
    \centering
    \caption{Table of optimal play calls and corresponding utilities for scenarios where the offensive team has a small deficit.}
\end{figure}

Interestingly enough, field goals are recommended a lot more in these scenarios, as it finds that it may not be worth running another play to get closer to scoring when the kicking distance almost guarantees three points. 

\subsubsection{Tie Game}
The following table shows the play call distribution for decisions made in late game scenarios with the game tied, defined as a score differential of exactly zero, where the next team to score would take a lead.

\begin{figure}[H]
    \makebox[\textwidth][c]{\includegraphics[width = 4.5in,height=3in]{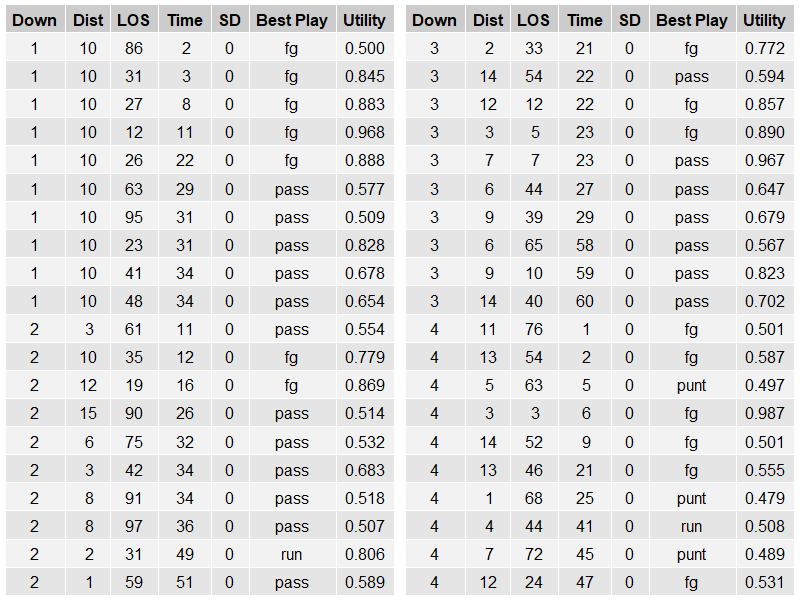}}
    \centering
    \caption{Table of optimal play calls and corresponding utilities for scenarios where the teams are tied.}
\end{figure}

We see that the recommendations tend to match those of the small deficit scenarios, opting to attempt to win in regulation if possible.

\subsection{Relationship Between Utilities and Expected Points}
The utilities derived for the non-late game scenarios have an interesting interpretation and relationship with the expected points model used widely in the NFL today. In the Burke (2009) original expected points model, the value was calculated using the ``average next score" approach, looking down the progression of the game for each play of a given scenario and averaging the points of the next scoring event. Recently, Yurko et al. (2018) has updated the expected point model to more directly model expected points in terms of the probability of each scoring event occurring given a set of preplay information. Using a multinomial logistic regression, the authors calculate the probability of each scoring event type and find the expected points by simply taking a weighted sum of the events with their corresponding probabilities. 

While this method makes sense and allows for a high level of intuition, in terms of helping a play caller make decisions for their team, we believe these methods can be improved. The model used in this paper is similar in that it looks into the future of a particular play, but there are three key differences in how our utilities are calculated from expected points. 

First, while real data is used to create the probability distributions for each play, the future possibilities used to create the final values for each play need not come directly from observed games, but rather are a probabilistically determined expected value of points given all of the possible routes one could take to the end of the drive. This allows for a more flexible yet more detailed approach to providing an expected points value, as it does not bias the data so strongly towards the observed results, and instead uses information that may not be directly from that specific scenario to make more informed estimates of the future results of a drive.

Second, instead of using the next score to determine the value to be used for each play, this method only uses the points scored at the end of the drive in the case that the scenario ends in a scoring event, and uses the expected utility the defense would get for taking possession of the ball at the yard line where the offense loses possession. This allows for the possessions to be relatively independent of one another, reducing the connection between distant future possessions while still including information about the future results to not ignore the connection between drives. It also continues the focus of a probabilistically complete evaluation of scenarios rather than relying on the limited results of previous drives.

Third, and possibly the largest deviation from previous methods, is that the utility values calculated here are those calculating assuming ``optimal decision making" occurs on future plays, where as the previous expected points calculations use the actual coaches decisions. To reiterate, this means that to calculate the utility value on first down, it assumes the coach will call the best play on the subsequent second down (or first down in the case where the first down is achieved on the first play), which again assumes the ideal action will be taken on third down. While this method does not optimistically assume that when the ``optimal" play call is chosen that the resulting play will always be successful, it does assume that on average, the results of choosing the ``optimal" play call will result in more points than choosing any other play at that point in the drive. 

Despite these differences, the utility calculation methods can still be viewed as an expected points model, as it still probabilistically calculates the expected value of each scenario of a game. Since the calculations are more robust in terms of using a full probability distribution for each play to determine the future results of the play, these values should provide more insight to decision makers when deciding what plays to call in each situation. While it is not required for any decision maker to perfectly follow the decisions made through these methods, any deviation from the recommended path would ultimately be at the cost of expected points, and therefore against what the historical data would indicate to be optimal. 

\begin{figure}[H]
    \makebox[\textwidth][c]{\includegraphics[width = 4.5in,height=3in]{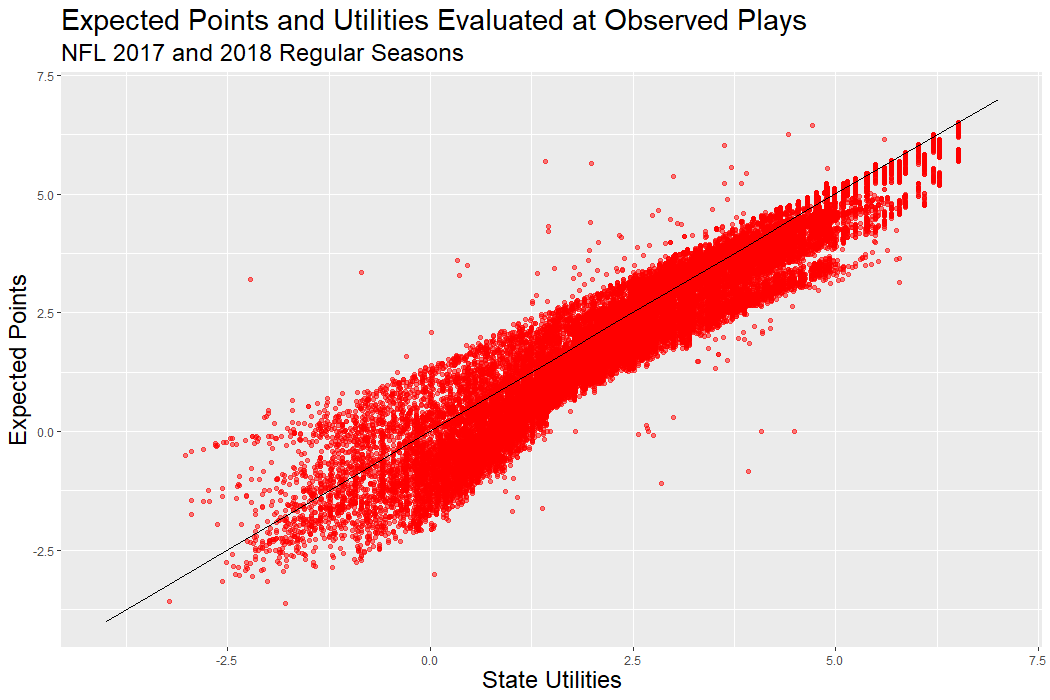}}
    \centering
    \caption{Expect points calculated via Yurko et al. (2018) versus state utilities.}
    \label{fig11}
\end{figure}

Given the connection, a good litmus test for our utility values would be whether or not they are correlated with the expected point values calculated using the previous methods. We would expect our values to be similar, however due to our values coming from ``optimal" decisions, we would expect our values to be on average greater than those provided, although not always due to the fluctuations in observed scores in the data that may drive expected point values higher. Fig.~\ref{fig11} shows the utility values plotted with their corresponding expected points values, indicating a high correlation of 0.95 between the methods along with the tendency for the utility values to be greater than the original expected points metric.

\subsection{Late Game Utilities as Win Probabilities}
Similar to the relationship between expected points and the non-situational utilities, there exists an analogous relationship between the situational utilities calculated and the win probability metric. From the same source that standardized the expected points metric, Yurko et al. (2018), win probability is calculated through a generalized additive model (GAM) to estimate a team's eventual probability of winning the game given a game situation including information such as expected score differential and time remaining. Using cross validation, the authors tuned their model to best calibrate their probabilities with observed games. 

Since our late game method assigns a utility value of one when a team would win with 100\% probability, and conversely a value of zero when the probability of a team winning drops to 0\%, our late game utilities can be considered another estimation of a team's win probability. These values are calculating by looking into the future of a team's current situation and calculating the scenarios in which they win or lose and returns a number that represents the proportion of those options in which they win, thus the win probability interpretation is reasonable. Similar to the differences in the previous set of metrics, our methods differ in that we use the full probability distributions and we assume perfect play calling for the offensive teams. However, even though our model implies optimal play calling for the offense, it also implies that a defense that obtains possession of the ball would also optimally call plays, therefore the positive drift created by optimal play calling should be lessened to an extent. We would expect that the likelihood of an offense scoring to be greater using the methods described in this paper, but the win probability to be on average the same as those found using previous methods.

\begin{figure}[H]
    \makebox[\textwidth][c]{\includegraphics[width = 4.5in,height=3in]{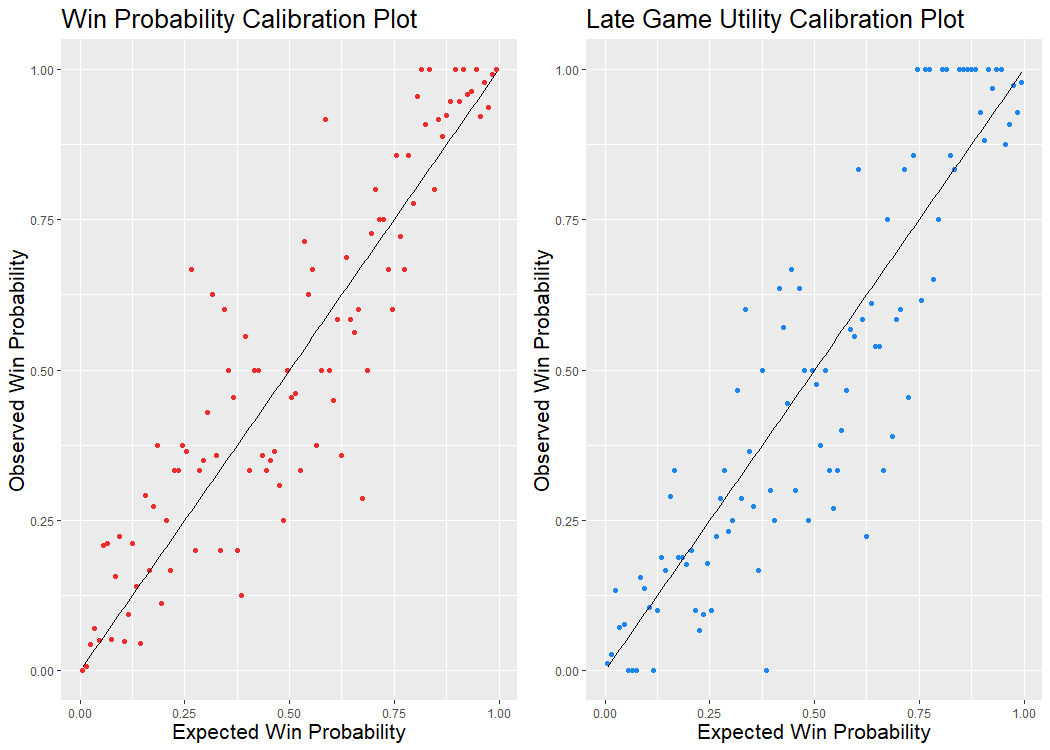}}
    \centering
    \caption{Calibration plots for two differing win probability metrics, with the previous method shown in the left plot and using the late game utilities on the right.}
    \label{fig12}
\end{figure}

If indeed our values can represent win probabilities, a metric for testing the values would be calibration. Dawid 1982 tells us a well calibrated model implies for any given win probability percentage $x$, if we look at all instances where a win probability of $x\%$ was predicted, then approximately $x\%$ of those games would result in a victory for the offensive team. Fig.~\ref{fig12} shows side by side calibration plots of the late game utility values generated for the actual data for the relevant late game scenarios, along with a calibration plot for the corresponding win probabilities calculated using previous methods.

The values have been put in bins with a width of $1\%$. If these methods were perfectly calibrated, the dots would be perfectly along the black line. It is clear that while this method does not show perfect calibration, it does follow the line relatively closely, implying that the model works to a reasonable degree. We see that in comparison to the previous win probability model, the late game utilities are comparable in their level of calibration, with possible exceptions occurring at the extreme ends. Our model indicates a tendency to be under-confident in predicting victory or defeat for a team near the end of the game. This is due to a multitude of factors, the most notable of which being the lack of scenarios to evaluate, the utility values not having reached convergence, and a higher probability allowed for the defensive team to gain possession and take the lead. 

\subsection{Evaluation of Teams as Decision Makers}
Using the metrics derived in this paper, we can now evaluate teams in terms of their decision making abilities. Assuming that our methods do capture the optimal decision making strategies, we can evaluate teams based of their percentage of play calls matching the optimal play call as decided by the reinforcement learning techniques. Thus, using the plays from the NFL 2019 regular season (not included in the data for training the utilities), we calculated the proportion of plays in which each team chose the correct play call in each scenario. We limited this evaluation to only plays in the game in which the team actually called a run or pass play, therefore not including many fourth down plays where a punt or field goal may have been chosen. 

\begin{figure}[H]
    \makebox[\textwidth][c]{\includegraphics[width = 4.5in,height=3in]{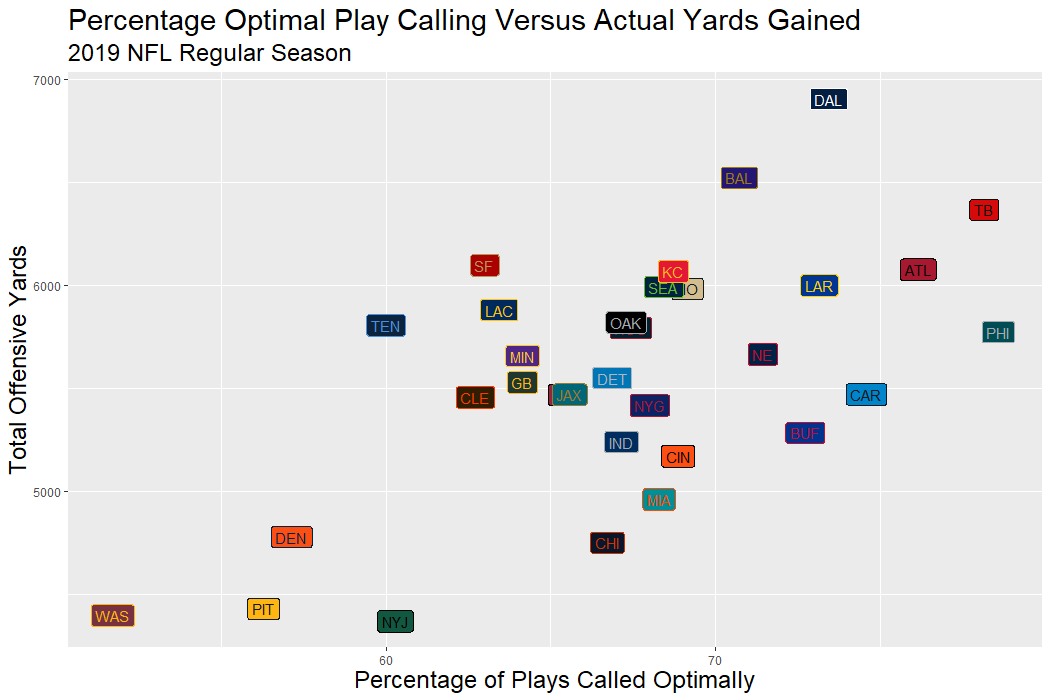}}
    \centering
    \caption{Percentage of plays called optimally plotted against total offense for each team.}
    \label{fig13}
\end{figure}

Fig.~\ref{fig13} shows each team's percentage of optimal play calls chosen plotted against their total offensive yardage. If the methods are accurate, then one would expect the teams that call a higher proportion of optimal plays would accumulate more yards on average than others. The positive trend in the graph tends to validate this conclusion. 

It is again worth noting that the ``optimal" play calls were determined using a collective league average offense, and therefore some teams may be calling plays more or less optimally based off their own offensive prowess. The differences may most likely be seen in run-heavy teams like Tennessee, San Francisco, and Baltimore, who would likely have more favorable probability distributions for run plays, implying there are plays when they ran the ball at an optimal moment when most teams would actually do better passing.

\subsection{Simulated Drives Against Historical Opponents}
Finally, if our methods are indeed superior to historical play callers, then we would expect our ``optimal" play calls to generate more points than a coach that calls plays according to historical trends. To test this, we selected 1000 different random combinations of $DOWN$, $DIST$, and $LOS$ and simulated games between play calling systems. It should be noted that the $DIST$ values were selected psuedo-randomly to limit the number of unlikely scenarios, limiting $DIST$ values to be equal to 10 if $DOWN = 1$ unless $LOS < 10$ (in the case of first and goal), and otherwise limiting $DIST$ to be less than 15.

First, the utility based play caller would be given the ball at a given $DOWN$, $DIST$, and $LOS$ and would determine the optimal play call. Then a result would be drawn from the necessary transition probability distribution, determining the subsequent state. In the new state the optimal play call would again be determined and a new state would be drawn accordingly. This would continue until the offense either entered a terminal state or gave the ball to the defense, at which point the historical play caller would have the opportunity to call plays from the resulting turnover position. The historical play caller would choose randomly from the set of actions based off the probability of each play type being called in the 2017 and 2018 regular seasons. This would repeat until one of the offenses (or defenses) scored points. This starting $DOWN$, $DIST$, and $LOS$ combination would be repeated 100 times, and the average score would be recorded. Then the historical play caller will be given the opportunity to start on offense from the same starting $DOWN$, $DIST$, and $LOS$ combination, following the same process for 100 new simulations. 

\begin{figure}[H]
    \makebox[\textwidth][c]{\includegraphics[width = 4.5in,height=3in]{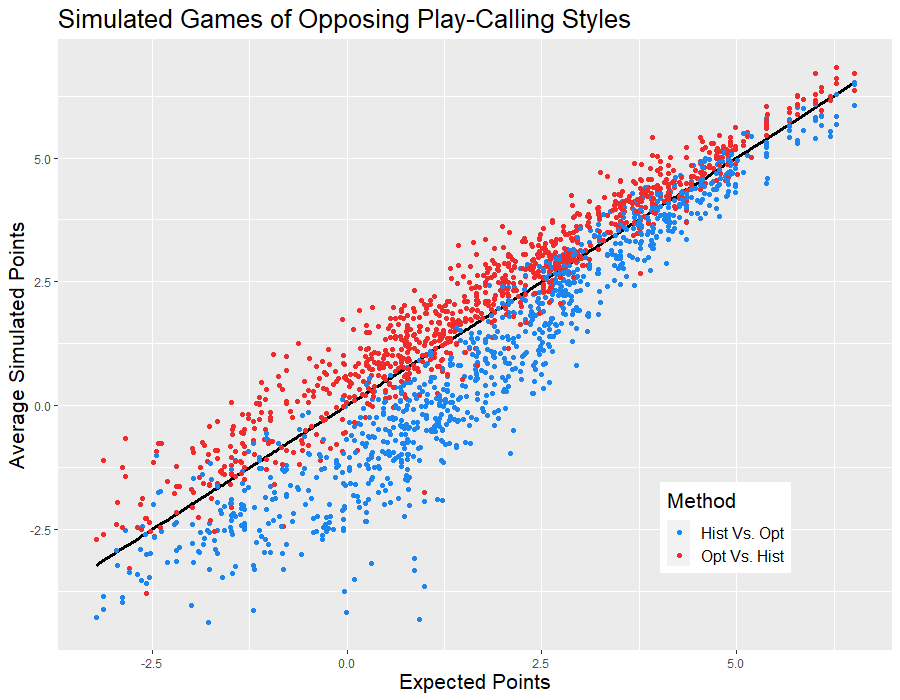}}
    \centering
    \caption{Simulated games between a utility driven play caller versus a historic play caller.}
    \label{SimGames}
\end{figure}

Fig. ~\ref{SimGames} shows the result of these simulations. Each red dot represents the average score of 100 simulated ``games" in which the utility based play caller was the initial offense for a particular $DOWN$, $DIST$, and $LOS$ combination. For each red point, there is a corresponding blue point for the same starting $DOWN$, $DIST$, and $LOS$ combination, representing the average score when the historical play caller starts on offense. The black line represents the state utilities for each scenario, marking the expected points an offense would obtain if both play callers made optimal decisions.

While there are certainly multiple cases in which the historic play caller manages to beat the utility driven decision maker, it is clear that the ``optimal" play caller is more successful on average. We see that most of the red points fall above the red line, implying that the offense tends to both make the right plays on offense as well as capitalize on their opponents mistakes, scoring more points than they would be expected to score given their opponent made optimal decisions. The gap in points scored between the play-calling systems leads us to believe that there is much room to grow in the decisions being made by coaches, and that a data driven approach could create a meaningful swing in production.

\section{Conclusions}
The results of this paper indicate two main points:
\begin{itemize}
    \item Coaches who do not use data to inform their decisions leave a lot of room for improvement, particularly in terms of offensive play calling, and
    \item The football analytics community needs to transition the evaluation of decisions and plays to a focus on probabilistic outcomes rather than the actual results.
\end{itemize}
Using the methods discussed here, we believe that coaches could see a strong increase in offensive productivity if data were properly used for play calling, assuming that proper caveats are considered for individual team strengths and opponent adjustments. While following these methods perfectly do not guarantee success at any level of football, the data suggests a gap in productivity that could be recouped by a savvy team of decision makers. 

Likewise, while previous methods of evaluation have generated insightful and meaningful results, they seem to have been over-reliant on the specific previously observed data and lacked the completeness offered by a reinforcement learning based approach. Adjusting the methods proposed by previous authors would not require a change in interpretation of previous results unless the data warranted a new understanding of the provided information. Adopting the methods discussed in this paper should provide novel valuable insights and allow for an improved comprehension of the game.

There are many ways this work can be expanded, some of which have already been discussed. Outside of improving the modeling methodology and relaxing some of the stricter assumptions, it would be helpful to expand the action space to include more than the two basic options. By doing this, a coach could know which play in their playbook is optimal in a given scenario outside of the generic suggestion to run or pass, allowing them a more direct path to exploit a defense. Additionally, it will be important to address the predictability of following an optimal play calling scheme. For example, if an opposing coach were to know that a team followed the recommendations from this system perfectly, can he coach his defense to stop the offense using the knowledge of which play is coming next? In other words, can a defense affect the transition probabilities enough to change the offense's optimal play calls? This is an open question and is worth exploring.

\section*{References}
\begin{enumerate}

    \item Bai, Y. and Jin, C. (2020). Provable self--play algorithms for competitive reinforcement learning. https://arxiv.org/pdf/2002.04017.pdf.

    \item Baldwin, B. (2018): ``Rushing Success and Play-Action Passing," \textit{Football Outsiders: Stat Analysis}, URL footballoutsiders.com/stat-analysis/2018/ru-shing-success-and-play-action-passing.
    
    \item Bellman, R.E. (1957). ``Dynamic Programming,'' Dover. 
    
    \item Boronico, J. and S. Newbert (2007): ``An empirically driven mathematical modelling analysis for play calling strategy in American football," \textit{European Sport Management Quarterly}, 1:1, 21-38.
    
    \item Brighenti, C. (2019): ``Modelling Run Outcomes by Field Control," \textit{Kaggle: 2020 NFL Big Data Bowl}.
    
    \item Burke, B. (2009): ``Expected Point Values," \textit{Advanced Football Analytics}, URL http://archive.advancedfootballanalytics.com/2009/12/expected-point-values.html.
    
    \item Burke, B. (2010): ``Fumble Rates by Play Type," \textit{Advanced Football Analytics}, URL http://archive.advancedfootballanalytics.com/2010/01/fumble-rates-by-play-type.
    
    \item Burke, B. (2014): ``Expected Points and Expected Points Added Explained," \textit{Advanced Football Analytics}, URL https://www.advancedfootball\\ analytics.com/index.php/home/stats/stats-explained/expected-points-and-epa-explained.
    
    \item Cafarelli, R., C. J. Rigdon, and S. E. Rigdon (2012): ``Models for Third Down Conversion in the National Football League," \textit{Journal of Quantitative Analysis in Sports}, 10.
    
    \item Carroll, B., P. Palmer, J. Thorn, and D. Pietrusza (1988): ``The Hidden Game of Football," New York, New York: Total Sports, Inc.
    
    \item Carter, V. and R. Machol (1971): ``Technical Note - Operations Research on Football," \textit{Operations Research}, 19, 541–544.
    
    \item Clement, S. (2018): ``Seahawks couldn’t ``establish the run” because there’s no such thing," \textit{SBNation: Field Gulls}, URL https://www.fieldgulls.com-/2018/1/3/16808842/seahawks-establish-the-run-myth-nfl-analytics.
    
    \item Daly-Grafstein, D. and L. Bornn (2019): ``Rao-Blackwellizing Field Goal Percentage," \textit{Journal of Quantitative Analysis in Sports}, 15.
    
    \item Dawid, A. (1982): ``The Well-Calibrated Bayesian," \textit{Journal of the American Statistical Association}, 77:379, 605--610.
    
    \item ESPN (2019): ``NFL Team Total Offense Regular Season Stats 2019," \textit{ESPN: NFL}, URL https://www.espn.com/nfl/stats/team/\_/season/2019-/seasontype/2.
    
    \item Goldner, K. (2017): ``Situational success: Evaluating decision-making in football," \textit{Handbook of Statistical Methods and Analyses in Sports}, 8.
    
    \item Hirshleifer, J. and J. Riley (1992): ``The Analytics of Uncertainty and Information," \textit{Cambridge Surveys of Economic Literature}.
    
    \item Ho, D., K. Imai, G. King, and E. Stuart (2011). ``MatchIt: Nonparametric Preprocessing for Parametric Causal Inference." \textit{Journal of Statistical Software}, Vol. 42, No. 8, pp. 1-28. URL https://www.jstatsoft.org/v42/i08/.
    
    \item Hopfield, J.J. and Tank, D.W. (1985). Neural computation of decisions in optimization problems. \textit{Biological Cybernetics} 52, 141--152.
    
    \item Horowitz, M., R. Yurko, and S. L. Ventura (2017): ``nflscrapR: Compiling the NFL play-by-play API for easy use in R," URL https://github.com/mak-simhorowitz/nflscrapR, r package version 1.4.0.
    
    \item Jordan, J., S. Melouk, and M. Perry (2009): ``Optimizing Football Game Play Calling," \textit{Journal of Quantitative Analysis in Sports}, 5.
    
    \item Leake, J. and N. Pritchard (2016): ``The Advantage of the Coin Toss for the New Overtime System in the National Football League." \textit{The College Mathematics Journal}, 47(1), 2-9.
    
    \item Lopez, M. (2014): ``Penalty rates in the NFL," \textit{StatsbyLopez}, URL https://-statsbylopez.com/2014/03/05/penalty-rates-in-the-nfl/.
    
    \item Lopez, M. (2017): ``Logistic regression and NFL kickers," \textit{StatsbyLopez}, URL https://statsbylopez.files.wordpress.com/2016/01/lab4.pdf.
    
    \item Morris, B. (2017): ``When To Go For 2, For Real," \textit{fivethirtyeight: Sports}, URL https://fivethirtyeight.com/features/when-to-go-for-2-for-real/.
    
    \item Pash, G. and W. Powell (2019): ``A Mixed-Data Predictive Model for the Success of Rush Attempts in the National Football League," \textit{Kaggle: 2020 NFL Big Data Bowl}.
    
    \item Pasteur, R. D. and K. Cunningham-Rhoads (2014): ``An expectation-based metric for NFL field goal kickers," \textit{Journal of Quantitative Analysis in Sports}, 10.
    
    \item Ploenzke, M. (2019): ``NFL Big Data Bowl Sub-Contest," \textit{Kaggle: 2020 NFL Big Data Bowl}.
    
    \item Rumsey, K. and B. DeFlon (2019): ``Quantifying Leverage at the Point of Attack," \textit{Kaggle: 2020 NFL Big Data Bowl}.
    
    \item Sahi, S. and M. Shubik (1988): ``A model of a sudden-death field-goal football game as a sequential duel," \textit{Mathematical Social Sciences}, 15.
    
    \item Silver, D., Schrittweiser, J. et al. (2017). Mastering the game of Go without human knowledge. Deep Mind Publication. https://discovery.ucl.ac.uk-/id/eprint/10045895/1/agz-unformatted-nature.pdf.
    
    \item Smith, M. (2019): "Onside kick success dropped from 21 percent to 6 percent after new rule," \textit{NBC Sports}, URL https://profootballtalk.nbcsports\\ .com/2019/11/03/onside-kick-success-dropped-from-21-percent-to-6-percent-after-new-rule/.
    
    \item Somers, T. (2020): ``Rantalytics: Does running the football successfully help the play-action pass?," \textit{USA Today Sports: Cowboys Wire}, URL https://cowboyswire.usatoday.com/2020/01/18/nfl-analytics-study-running-the-ball-play-action-success-mike-mccarthy-dallas-cowboys/.
    
    \item Stern, A. (2019): ``Practical Applications of Space Creation for the Modern NFL Franchise," \textit{Kaggle: 2020 NFL Big Data Bowl}.
    
    \item Sutton, R. and A. Barto (2018): ``Reinforcement Learning: An Introduction," \textit{MIT Press}, Cambridge, Massachusetts.
    
    \item White, C.C. and White, D.J. (1989). ``Markov decision processes,'' \textit{European Journal of Operational Research} 39, 1--16.
    
    \item Woodroofe, M. (1979). ``A one--armed bandit problem with a concomitant variable,'' \textit{Journal of the American Statistical Association} 74, 799--806.
    
    \item Yam, D. and M. Lopez (2019): ``What Was Lost? A Causal Estimate of Fourth down Behavior in the National Football League," \textit{Journal of Sports Analytics}, 5:3, 153-167.

    \item Young, C. (2020): ``Applying Machine Learning to Predict BYU Football Play Success," \textit{Towards Data Science}, URL https://towardsdatascience.com\\
    /applying-machine-learning-to-predict-byu-football-play-success-60b57267b78c.
    
    \item Yurko, R., S. Ventura and M. Horowitz (2019): ``nflWAR: a reproduceible method for offensive player evaluation in football," \textit{Journal of Quantitative Analysis in Sports}, 15.
    
    \item Zauzmer, B. (2014): ``Modeling NFL Overtime as a Markov Chain," \textit{Harvard Sports Analysis Collective}, URL http://harvardsportsanalysis.org/20-14/01/modeling-nfl-overtime-as-a-markov-chain/.
\end{enumerate}

\newpage


\section*{Appendix A: Data}
The data used was the play--by--play data for all games from the 2017 and 2018 regular seasons. This data can be found through the nflscrapR package developed in Horowitz et al. (2017). Additionally, data was obtained through ESPN for offensive yards in the 2019 season.

\subsection*{A.1 Acquiring data for specific plays}
Part of the task of creating probability distributions requires finding a subset of the overall set of observed plays that should be considered relevant for the play in question. Unfortunately, due to data sparsity, this is much more difficult than finding exact matches. For example, while 1st and 10 from the 40 yard line occurred over 350 times in the two seasons examined, looking at 2nd and 10 from the same yard line drops to under 70 occurrences, down to only 15 times for 3rd down, and only four times for 4th down (all of which were punts). Considering that we need to model each of these plays for both run and pass events, it will be extremely difficult to achieve any level of precision in estimation if we limit our data to only the exact matches.

Therefore, to acquire data for each play scenario, we use the Ho et al. (2011) MatchIt R package to find similar sets of plays. Using the Mahalanobis distance, we find the 100 most closely related data points for a specified set of base preplay information, and use the observed yards gained from the plays to create a probability distribution for the play scenario. This method would allow for a quick subset of the observed data to be found that matches closely in terms of $DIST$ and $LOS$, with the value of the $DOWN$ almost always being exactly matched. Exceptions occurred on fourth downs, where very few non-kicking plays were observed, and therefore data was often borrowed from third downs as decided by the matching process. The number of 100 data points was specifically picked after examining several different cutoff values and finding it to be a value that both created consistently smooth and accurate distributions while not allowing the flexibility to become too liberal to include data unrelated to the original play. 

Some slight modifications had to be made to retain the context of the observed plays intact. For example, when we are trying to model a pass play on 3rd and 10, often we will pull information from similar plays, such as 3rd and 8. However, on 3rd and 8, a gain of 8 yards would result in a first down, and therefore be considered a success. However, if we simply transfer the 8 yard gain into the 3rd and 10 yard context, an 8 yard gain would be considered falling short, and therefore a failure. Conversely, in a 3rd and 12 situation, gaining 10 yards would be considered a failure, but if that were transferred directly to the 3rd and 10 scenario, we would have a success. To keep the consistency of ``success" or ``failure" on a play, we will scale each observed yards gained from matched plays by $DIST_{actual}/DIST_{borrow}$, where $DIST_{actual}$ is the distance to the first down marker on the desired play, and $DIST_{borrow}$ is the distance to the first down marker on the play being used to estimate the distribution. For example, on the 3rd and 10 play, if we borrowed an 9 yard gain from a 3rd and 8 play, the 9 yard gain would be scaled up by $10/8$, making it effectively a 11.25 yard gain. A max scale of 1.5 was used, limiting the adjustment of yards gained in plays borrowed to not inflate moderate gains in short yardage situations. Additionally, cutoffs were made for large gains (and losses) at the boundaries to ensure no play could be made for a gain of more yards than were physically possible (i.e. $LOS - 100 \leq GAIN \leq LOS$). 

\section*{Appendix B: Modeling} 
Determining the probability distributions of each type of play had to be done carefully in order to ensure the statistical stability of the decision making process. In previous studies, usually evaluations are made using the pure ``what happened next" approach that does not explore the totality of options that could occur on any given play. In addition to having a limited number of observations available for each scenario, using exact state data also does not allow for one to incorporate a priori knowledge of the game of football.  Therefore, using just the previously observed situations for these events will be biased towards the events that occurred, and therefore a more comprehensive method of modeling the data would be preferred. 

To do this, we use different methods to evaluate run and pass plays. Run plays tend to have a very smooth curve, having a heavy-tailed asymmetric distribution, truncated at both ends by the end zones. Pass plays tend to have a mixed distribution, with a discrete point mass at the line of scrimmage coming from incomplete passes, and a continuous multimodal distribution for the other plays, with peaks occurring slightly behind the line of scrimmage (usually from sacks) and slightly past the line of scrimmage and slowly decreasing from there. The following subsections detail the assumptions and modeling process for each of the actions, including the less common actions.

\subsection*{B.1 Run Plays}
On a typical run play, yards gained follows a generally smooth distribution, peaking around two to five yards gained, depending on the specific down, distance, and distance to the end zone. The distribution tails off quickly to the negative side and slowly to the positive side of the mode, indicating it is more likely for a team to gain a large amount of yards than lose a large amount of yards. This is in keeping with our understanding of the game, as it is more likely to observe a 15 yard run than it is a -5 yard run. Finally, near both end zones, we observe a truncation of the distribution, where the max value of yards gained is limited either by the distance to scoring a touchdown or being tackled behind the line of scrimmage for a safety.

\begin{figure}[H]
    \makebox[\textwidth][c]{\includegraphics[width = 4.5in,height=3in]{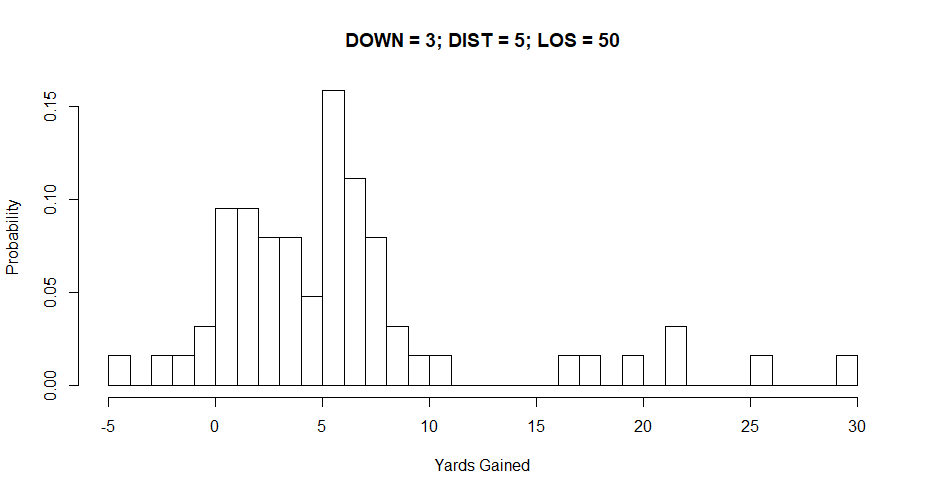}}
    \centering
    \caption{Distribution of yards gained on run plays for 3rd and 5 at midfield.}
    \label{RunDist}
\end{figure}

One observation about the distribution of run plays that was not particularly expected was the consistency of smaller modes further along the distribution. While this phenomenon could be chalked up to small sample sizes, there are also contextual factors specific to the game of football that could explain their regularity. For example, in Fig.~\ref{RunDist}, we observe the results of run plays that were called on 3rd and 5 at midfield.
In this histogram, we actually observe three areas that could be interpreted potentially as modes. The largest occurs just past five yards gained, which would imply the runner gained just enough for a first down. The other two occur around two yards and twenty yards gained, where the first would suggest the average yards gained of a defense that was prepared to stop the run, and the latter would suggest a defense that was completely unprepared to stop the run and the runner broke free for a large gain. This could also be perceived as the different ``lines of defense" prepared to stop the run. On a moderate defensive stop, typically the runner is being tackled closer to the line of scrimmage often by a defensive lineman or linebacker. For a moderate offensive gain, the first line of defense has typically been broken, and the tackle must be made further down the field by a linebacker or safety. On a strong offensive gain, typically the tackle is occurring even further down the field, meaning the designed defense has likely busted, and the tackle is being made by a safety or cornerback, or another player who was not in position to make the play earlier.

This similar structure can be observed in several running scenarios: a group of results that indicate a moderate defensive stand, a group of results that indicate moderate offensive success, and a group of results that indicate the breaking free of the runner for a large gain. Additionally, we often will observe a fourth group representing a strong defensive stand, resulting in a group of yardage gained values less than zero. As one would expect, the two moderate groups would have more probable modes than the more extreme groups, as gains of more than 10 yards or being tackled for a loss on a run play tend to be rather rare by comparison. The relative size of the first two modes typically depends on the down and distance and whether or not a run play was anticipated by the defense, therefore the average gains tend to reflect an inverse relationship with the play's distinction of a ``running down" (a down where a coach would often suggest running the ball, such as a short yardage situation or early down).

This does not necessarily suggest that running the ball is more successful on ``passing downs" (opposite of running down, typically later downs or those with large $DIST$ values), but simply that the average yards gained typically is higher when the defense is prepared to stop a different type of play. This may seem paradoxical, but it makes sense in context. For example, a run play on 3rd and 5 from the 50 yard line (typically considered a passing down) averages 6.40 yards, while a run play on 3rd and 1 from the 50 yard line (typically a running down) averages only 4.18 yards. However, while the average yards gained is greater on the first, the probability of gaining a first down on the first play is only 57\%, while the probability of gaining a first down on the second play is around 83\%. If we compare these probabilities to their respective first down probabilities for choosing to pass, we observe that the first play achieves a first down via pass about 38\% of the time, while the second play achieves a first down only 43\% of the time. Thus the gain in achieving a first down for running in this short yardage context is about 40\%, while the gain of achieving a first down by running on the midrange play is only about 20\%. Thus, while running in longer distance scenarios may achieve more yards on average due to the response of the defense, it still may not achieve better success in leading to points. Note that all of the probabilities listed here are actual observed likelihoods, not modeled values. 

To address the idea of multiple modes within run play distributions, we choose to model all run plays with a mixture model with four normal distributions. These are to allow for the potential presence of each of the four groups (moderate offensive run, moderate defensive stop, strong offensive gain, strong defensive stop). However, we allow the data to determine the location and relative weight of each of these distributions, effectively allowing one or more of the distributions to go to zero if the data does not indicate evidence of a particular group on a particular play. Thus, we model the run plays as hierarchically follows, training the parameter values via Gibbs sampling.
$$    y_{run} \sim \sum\limits_{k=1}^4 w_k \mbox{N}(\mu_k,\sigma^2),\quad
\sum\limits_{k=1}^4 w_k = 1,\,\,w_k\geq 0,
$$
with priors $\sigma^2\sim IG(1,1)$, an inverse gamma distribution, $\mu_k\sim \mbox{N}(0,100)$, and $w\sim Dir(1,1,1,1)$, a Dirichlet distribution.

\subsection*{B.2 Pass Plays}
Pass plays follow a less smooth distribution than run plays for obvious reasons. The biggest distinction is that pass plays have a mixed continuous-discrete density function due to the nature of the incomplete pass. As the saying goes, there are three things that can happen when the ball is thrown, two of which are bad:
\begin{itemize}
    \item The pass is caught by an offensive player (completed pass),
    \item The pass is caught by a defensive player (interception), or
    \item The pass is not caught (incomplete pass).
\end{itemize}
We will leave the discussion of the interception (as well as other turnovers) for Section B.6. When a complete pass event occurs, the whole range of yards gained outcomes are available, spanning from scoring a touchdown (gaining yards equal to the distance from the end zone, $LOS$), to being tackled in the offensive team's end zone for a safety. Thus, for completed passes, we would expect there to exist some level of a smooth curve. However, in the event of an incomplete pass, the result is always the same: a gain of zero yards. The ball is placed back at the original line of scrimmage at the start of the play, and the offense maintains possession given it was not fourth down, in which case the defense would take over possession at that yard line. Thus, the probability distribution should be expected to be rather smooth, with a point mass occurring at zero yards gained.

\begin{figure}[H]
    \makebox[\textwidth][c]{\includegraphics[width = 4.5in,height=3in]{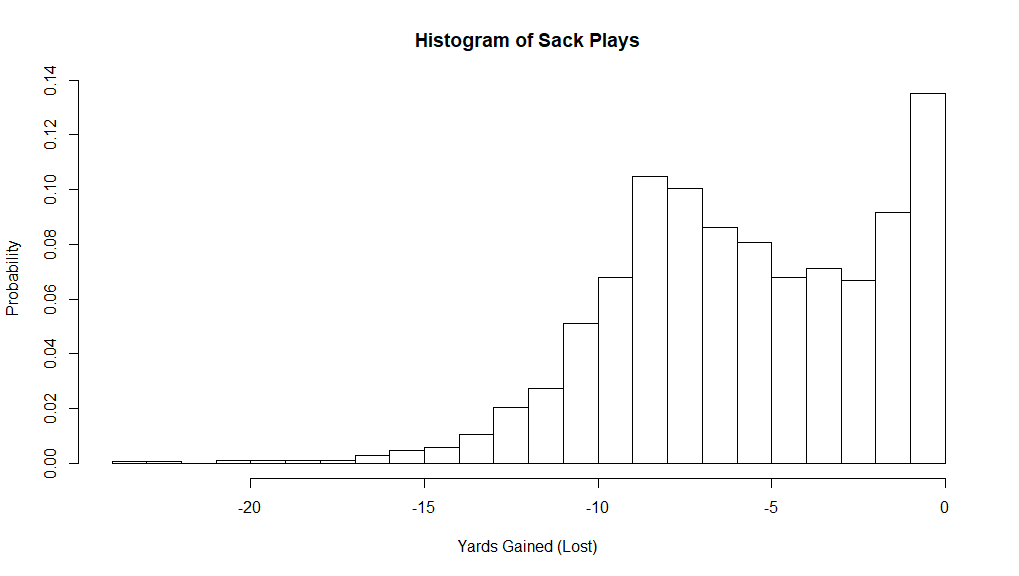}}
    \centering
    \caption{Distribution of yards lost on pass plays ending behind the line of scrimmage, usually resulting from sacks.}
    \label{SackDist}
\end{figure}

However, there is a slight complication in pass plays that result in a slightly less smooth curve. While quarterbacks intend to throw the ball on every pass play, often the defense is able to apply pressure to the point of achieving a sack. In that case, the play is still marked as a pass play, but the yardage gained is always less than zero. Because of this, and due to the location of the ``pocket" (area in which the quarterback stands when preparing where to throw the ball), the distribution of pass plays that end behind the line of scrimmage is rather hard to model with a parametric distribution. Fig.~\ref{SackDist} shows the data from these plays.

Additionally, the data indicate that pass plays tend to have distinct point masses of probability occurring at the touchdown, meaning that gaining enough yards to score a touchdown on a pass play is not necessarily an outlier, even for large values of $LOS$. While run plays also have the ability for players to break free of all the defenders to score on a long run, pass plays seem to more consistently result in scoring, to the point that we believe the touchdown pass should be expressly included in our model as a small point mass. This feature likely results from the fact that most deep pass plays that are completed would only require an offensive player to avoid being tackled by at most three defenders, and would likely already be moving at a high speed, and potentially would have created space between himself and the defenders given that he caught the pass. Thus, these combined features tend to result in more touchdowns than run plays, and therefore we will treat them separately from other completed passes.

It's also worth noting that there is one more event that can occur on a pass play, which is the case of the quarterback scrambling for positive yardage. In this scenario, the quarterback either senses pressure in the pocket or sees an opportunity to gain yards with his legs and opts to run the ball. In the case that he succeeds and is not tackled at or behind the line of scrimmage (classified as a sack), the play actually ends up being labeled as a run play. This causes difficulty in our data, as it assigns the yardage gained on what was designed to be a pass play to a running play, potentially creating bias in both distributions. Unfortunately, there is no simple method of resolving this issue without massive data augmentation. However, it should be noted and considered when discussing the utility of each play type that run plays will include the successful scramble plays and therefore a designed pass play may have greater utility than is determined using the data in this research.

Nevertheless, we must proceed in modeling pass plays. To deal with all of the features discussed, we will choose to model the yards gained on a pass play using a mixed distribution, dividing each of the potential results on a play to a specific case. On a pass play, we can classify the results of the play as one of the following:
\begin{itemize}
    \item An incomplete pass;
    \item A completed pass for a positive gain, not resulting in a touchdown;
    \item A completed pass for a negative gain (loss), or a sack resulting in a loss of yards;
    \item A completed pass resulting in a touchdown;
    \item A turnover, via either a fumble or an interception.
\end{itemize}
Thus, we have multiple continuous distributions and multiple point masses for yards gained. We will treat the turnover scenario as a point mass for now and discuss how it is divided into the resulting yard lines later. Using this, we can model individual plays by first treating the data as if it were sampled first from a categorical distribution, choosing one of the five scenarios described above, and then sampling from the relevant continuous distribution in the case of a completed pass or sack. Thus we model the pass plays as follows:
\begin{align*}
    z_{pass\_result} \sim 
    \begin{cases}
    INC \hspace{.3in} &w.p. \hspace{.1in} p_{INC}\\
    TD \hspace{.3in} &w.p. \hspace{.1in} p_{TD}\\
    POS \hspace{.3in} &w.p. \hspace{.1in} p_{POS}\\
    NEG \hspace{.3in} &w.p. \hspace{.1in} p_{NEG}\\
    TO \hspace{.3in} &w.p. \hspace{.1in} p_{TO}\\
    \end{cases}
\end{align*}
with
    $[y_{pass}|(z_{pass\_result} = INC)] = 0 $, 
    $[y_{pass}|(z_{pass\_result} = TD)]= LOS$,
    $[y_{pass}|(z_{pass\_result} = POS)] \sim f_{POS}$,
    $[y_{pass}|(z_{pass\_result} = NEG)] \sim f_{NEG}$,
    $[y_{pass}|(z_{pass\_result} = TO)] \sim f_{TO}$ and
\begin{align*}
    \textbf{p} &:= \{p_{INC},p_{TD},p_{POS},p_{NEG},p_{TO}\} \\
    &\sim Dir\bigg(10*\rho_{INC},10*\rho_{TD},10*\rho_{POS},10*\rho_{NEG},10*\rho_{TO}\bigg)
\end{align*}
The $\rho$ values are empirically computed from the data by calculating the probability of each of the five types of plays occurring given the line of scrimmage. This allows for an adjustable prior to be placed on the data to influence the categorical variable $z_{pass\_result}$ to look similar to that of the actual data, as each of the categorical events are not equally likely to happen given the yard line. For example, if a team is on the one yard line ($LOS = 1$), it is impossible for them to gain yardage but not score a touchdown, therefore the event $z_{pass\_result}$ should never be $POS$, and thus the value of $\rho_{POS} = 0$ for this case. In general, we see $\rho_{POS}$ increase as $LOS$ increases, and the opposite effect occur for $\rho_{NEG}$. We also see $\rho_{TD}$ increase as $LOS$ decreases (more likely to score a touchdown the closer a team is to the end zone). The values $\rho_{TO}$ and $\rho_{INC}$ are relatively stable, however, we do see slight increases of both values at both extremes for $LOS$. We multiply each of these $\rho$ values by 10 to give the prior a higher weighting, to imply that if we had no data to represent the play to be modeled, our posterior would still have a base of 10 plays to represent the information. Thus, as the number of plays used increases, this information becomes less and less meaningful. This combined with the data representing each play event creates a Dirichlet-Multinomial conjugate model from which to represent $z_{pass\_result}$.

From here, the task remains to define $f_{POS}$, $f_{NEG}$, and $f_{TO}$. First addressing $f_{NEG}$, the distribution of pass plays ending behind the line of scrimmage, we are mostly modeling sacks, although a fair portion of these plays are actual catches made behind the line of scrimmage. After looking extensively at these plays, we found there was no particular smooth structure in the negative distributions, and in fact what appeared to be oddities and outliers in the yards lost seemed to appear in a consistently noisy fashion. Therefore instead of modeling the negative distribution with a smooth function, we opted to use a Dirichlet-Multinomial conjugate distribution, treating the yards lost as one would a categorical variable with a fixed prior percentage of occurring, corresponding to the distribution seen in all negative plays. The likelihood would then provide a weighting observed in the specific plays pertaining to the specific base preplay information, and thus would create the full distribution.

\begin{figure}[H]
    \makebox[\textwidth][c]{\includegraphics[width = 4.5in,height=3in]{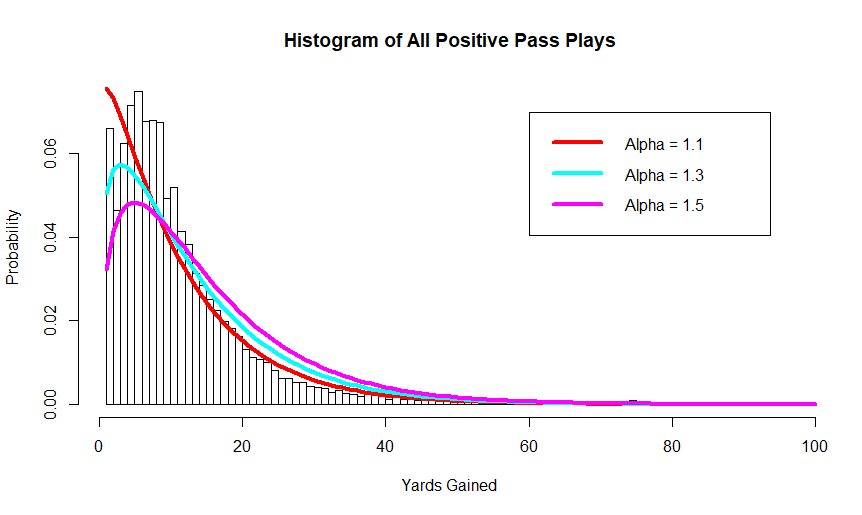}}
    \centering
    \caption{Histogram of yards gained on positive pass plays, with three suggested values of shape parameter $\alpha$.}
    \label{PosPass}
\end{figure}

To define $f_{POS}$, the distribution of yards gained given the pass was completed for positive yardage but not resulting in a turnover, we began by using the mixture of normal's model used for run play modeling. However, we found that the results were not typically satisfying, with the model exaggerating found modes in the yardage gained for higher values, upwardly biasing our distribution. Thus, an alternative method of using a Gamma-Gamma conjugate distribution to create the posterior. In this method, we assume the shape parameter, $\alpha$, is known and attempt to learn the rate parameter $\beta$. We determine the value of $\alpha$ by examining all pass plays as a whole and looking at the shape the distributions. From here, using knowledge of gamma distributions and how the we expect the shape and beta parameters to relate, we chose $\alpha$ to be equal to 1.3. Fig.~\ref{PosPass} illustrates evidence towards our conclusion.

As can be observed in the plot, an $\alpha$ value too close to one tends the mode toward zero, which we know not to be uncharacteristic of positive pass plays. However, an $\alpha$ value too large creates heavy tails and thus upwardly biases our posterior. Our chosen value of $\alpha$ of 1.3 allows us to meet in the middle, expecting that the error observed will be mitigated once the posterior is fit with observed data. From here, we choose the values of our prior parameters on our Gamma distribution, $\alpha_0$ and $\beta_0$, which we choose to be 130 and 1300 respectively, giving the interpretation of observing 100 pass plays that yield 1300 yards in total, which is reasonable. 

This process of selecting priors and a known parameter can be considered a loose version of Empirical Bayes, using knowledge of the game as well as the data to select reasonable shape and rate hyperparameters. While there are many methods available that would better fit the data, we wish to avoid using them for two reasons. One is because we believe the observed data is collected in a biased manner. Assuming that coaches are indeed good decision makers, they will in general have some level of intuition on to the relative expected result of a play. Therefore the data observed is a result from what coaches believe to be ideal play calls, and therefore not a truly random sample of the play calling space. In addition, there are many play calling scenarios in which we have very little observed data with which to model. In these cases, we will be fitting a distribution that does not have the proper amount of information to provide insight on the totality of outcomes. Thus, using a more advanced inference method to fit our parameter values would likely overvalue observed results and not properly identify the potential results of the play. Thus, we wish to use a predetermined prior to incorporate a limited amount of external knowledge of the game, while using the actual observed data to fill out the remainder of the distribution.

We can finish our modeling formulation with the following:
\begin{align*}
    f_{POS}|Y &\sim Ga\left(\alpha_0 + n * \alpha, \beta_0 + \sum y_{POS}\right)\\
    f_{NEG}|Y &\sim Dir\left(10 * \tau_{-1} + \sum \mathds{1}\left(y_{NEG} = -1\right), 10 * \tau_{-2} + \sum \mathds{1}\left(y_{NEG} = -2\right)\right.,\\
    &  \left.  \dots, 10 * \tau_{-24} + \sum \mathds{1}\left(y_{NEG} = -24\right)\right),
\end{align*}
where $\tau_{-k}$ is defined as the empirically determined probability of observing a loss of $k$ yards on any pass play, preprocessed and held constant for all pass plays. Also, $\mathds{1}\left(y_{NEG} = k\right)$ is defined as the indicator function evaluated as 1 when a pass play occurs for exactly $k$ yards. The weighting of 10 on each of the $\tau$ values can be perceived as weighting the prior term by 10, acting as 10 preobserved plays to limit the bias that could occur from individual plays having very few observed negative pass plays.

The modeling of the probability distributions for pass and run plays brings in the highest opportunities for error, and therefore the majority of error in the final results of this paper likely will be due to inaccurate modelling. By the very nature of this data, there are many scenarios that we are required to model that often do not have a large amount of existing plays with which to fit the model, requiring assumptions to be made that often lead to imprecise results. Thus, the biggest room for improvement in these methods lies in this task. There have been other methods of modeling run and pass plays (Young 2020, Ploenzke 2019, Rumsey and DeFlon 2019, Pash and Powell 2019, Stern 2019, Brighenti 2019), however, they either require outside sources of data to be used or are too restrictive to allow for all play scenarios to be modeled. However, much time was spent to examine the validity of the modeling decisions made here, and the level of error seemed to be minimal and not consistently biased in either direction. Thus, we will proceed forward assuming the distributions created are true, but leave the question open for how to improve upon these methods.

\subsection*{B.3 Punts}
\begin{figure}[H]
    \makebox[\textwidth][c]{\includegraphics[width = 4.5in,height=3in]{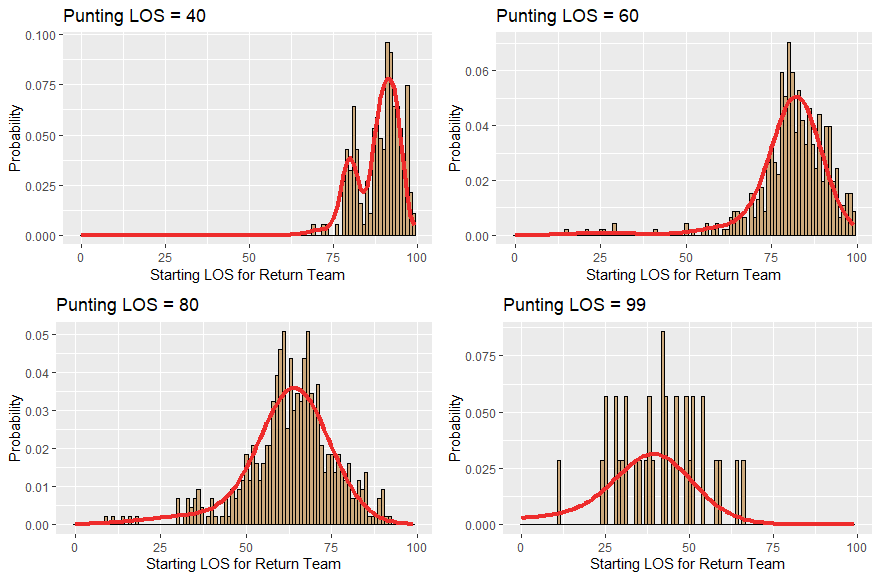}}
    \centering
    \caption{Distributions of net yards for punts at differing lines of scrimmage.}
    \label{PuntDist}
\end{figure}

Calculating the distribution of punt distances requires only one factor: the line of scrimmage at which the play started. By examining the punt distance trends for each line of scrimmage, it was observed that punts follow relatively smooth distributions, having a large mode about 40 to 50 yards from the $LOS$ and tapering off fairly smoothly in both directions. The exception to this occurs once the punting team gets closer to the offensive end zone ($LOS$ decreasing) and has to start adjusting the punt distance to avoid touchbacks. In either case, we modeled the punt distance distributions again using a mixture of normals model, as it seemed to fit both the pure unimodal distributions found in midfield punts and adjust well to the abnormalities that come into play closer to the end zones. No adjustments needed to be made for point masses for scoring touchdowns or touchbacks, as the data did not seem to indicate any significant increase in probability of those two events occurring compared to their adjacent results. Fig.~\ref{PuntDist} shows a few examples of the punt distributions modeled in red over their corresponding observed distances. Note that the distribution represents the net result of the punt in terms of the receiving team's (future offense) $LOS$ at the end of the play.

It should be mentioned that a few adjustments were made for completeness of data. To ensure both a smooth distribution and that enough data was used for each punt, when a $LOS$ value was selected, actual punt distances were pulled not only from the exact LOS value, but for all values within two yards of $LOS$. For example, for a punt happening at $LOS = 60$, we used all actual punt distances between $LOS = 58$ and $LOS = 62$, assuming that the distribution of punts would be relatively similar in a small neighborhood around $LOS$. Additionally, for punts occurring for $LOS$ less than 40, we assumed they followed the same distribution as that of $LOS = 40$. This was done due to the minimal amount of punts occurring from within 40 yards of scoring a touchdown, as most teams tended to opt to attempt a field goal or go for it on 4th down.

\subsection*{B.4 Field Goals}
Field goals has been modeled many times in many different formats (Daly-Grafstein and Bornn 2019, Lopez 2017, Pasteur and Cunningham-Rhoads 2010). For our purposes, we choose to follow the methods described in Lopez (2017), using a logistic regression using the distance of the kick as the only input feature. While many other factors such as wind speed, location of game (both as home/road teams, and locations where altitude may play a role), and even the individual kicker are obvious contributing factors, we choose to ignore them and continue with our method of using only the base preplay information that we feel is relevant, which can be simplified to just $LOS$ in this case. In the future, a team or organization could tailor this model to use the other pieces of information to potentially improve their results.

\begin{figure}[H]
    \makebox[\textwidth][c]{\includegraphics[width = 4.5in,height=3in]{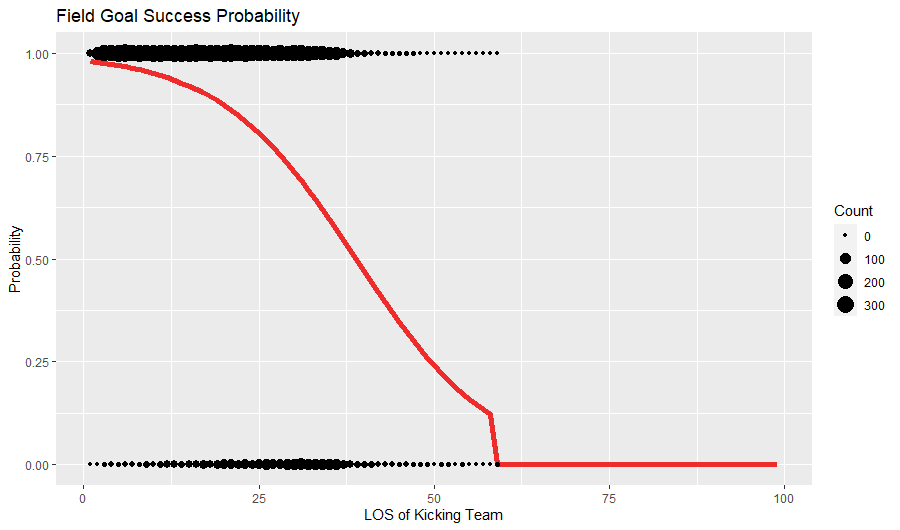}}
    \centering
    \caption{Probability of making a field goal at each different value of $LOS$, with the red line showing the modeled probability, and black dots indicating observed results.}
    \label{FGDist}
\end{figure}

The totality of modeling field goals therefore simplifies down to looking at the future states given the field goal is made or missed. The third but less common option would be the scenario of a blocked field goal (or even less so, a field goal coming up short of the goalposts and being returned by the defending team). However, given the rarity of these scenarios, compounded by the additional rarity of a blocked field goal being returned (as it is often more beneficial for the blocking team to leave the ball on the ground), we treat all cases of a failed field goal attempt as the same, with the future state being the defense gaining possession of the ball at the $LOS$ where the offense attempted the field goal. Thus the two future state options would be the terminal offensive field goal state, or the semi-terminal defensive taking possession of the ball state at $LOS_{k+1} = 100 - LOS_k$, which is the original line of scrimmage but in the point of view of the defense taking control offensively. The probability of each of these events happening is simply the probability of the field goal being made, and one minus that probability, respectfully. Fig.~\ref{FGDist} shows the probabilities used for each yard line. There is a jump at $LOS = 59$, at which the data has no points of which a field goal was attempted, and therefore the probability ceases to be modeled via a logistic regression and is assumed to be zero. 

\subsection*{B.5 Kickoffs}
\begin{figure}[H]
    \makebox[\textwidth][c]{\includegraphics[width = 4.5in,height=3in]{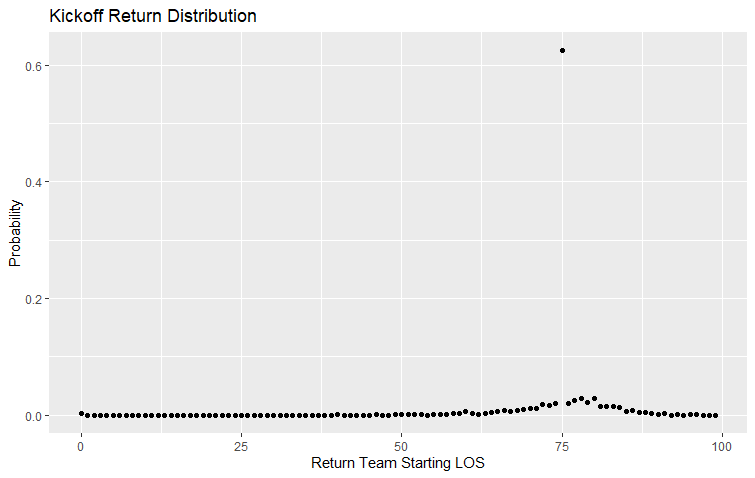}}
    \centering
    \caption{Distribution of starting $LOS$ values after kickoffs.}
    \label{KickoffDist}
\end{figure}

Kickoffs are a special scenario that occur at the beginning of each half and after each scoring play. There is never a choice to be made on whether or not a team is to kickoff. They are simply fixtures within the game that must be executed before proceeding to normal scrimmage plays. Luckily, there is a multitude of kickoff plays that have been completed, nearly all of which initiate from the same line of scrimmage (exceptions resulting only from penalties). Thus, rather than explicitly modeling kickoffs as seen for other play types, we will simply use the observed relative frequencies as the true probabilities. This results in a somewhat noisy but overall relatively smooth distribution, pictured in Fig.~\ref{KickoffDist}. Notice the spike occurring at $LOS = 75$, which corresponds to the touchback $LOS$.\\

Onside kicks, while very exciting when they work, are both rarely attempted and even less often recovered. Thus, we chose to leave them out of the calculations in this work, although they are admittedly a meaningful part of any end game scenario and should be considered for a more detailed approach. In order to add these into the system, one could simply opt to set a predetermined onside recovery value, perhaps the 2018-2019 average recovery rate of $6\%$ (Smith, 2019), and assume the recovery $LOS$ would be the same for the kicking and receiving team. Given the utilities should be similar for first and ten values within a small area of potential recoveries, this approximation of recovery location shouldn't create issues. Regardless, without doing an extensive analysis of onside kicks, their recovery percentages, and their resulting $LOS$ values, it feels unwise to include their effects within this research, thus it was not included.

It should be noted that the free kick event that results after a defense forces a safety is not technically a kickoff. Free kicks initiate from a different line of scrimmage than kickoffs (kicking team kicks from their own 20, rather than their 35), and therefore would not have an identical return distribution to that of kickoffs (although it might be quite similar). For the purposes of this research, the free kick was not modeled separately, and free kicks will be considered the same as kickoffs, despite their obvious differences. 

\subsection*{B.6 Dealing with Turnovers}
Turnovers can be split between two obvious cases: fumbles and interceptions. Interceptions occur when the quarterback (or other legal forward pass thrower) throws the ball forward and has the ball caught by a defensive player, and therefore  can only occur on pass plays. Fumbles occur on all other cases where an offensive player loses control of the football, and thus can occur on either run or pass plays. Additionally, interceptions always result in the defense gaining possession of the ball (outside of the rare circumstance when the defending player fumbles after creating the interception, which we are ignoring), however fumbles can be recovered by either the offense or defense. In our case, we care only of the cases where the fumble is recovered by the defense, as the alternative would not be considered a turnover and therefore can be included in normal offensive play results. The last complication of these events is there is no standard for how the ball is returned after either turnover event occurs, and thus requires us to model the probability of each starting $LOS$ after a fumble or interception event.

Examining plays for fumble and interception events individually helps identify how each play can be modeled. In general, the model has three noticeable features, common to both turnover types:
\begin{itemize}
    \item the data is unimodal, centered about the $LOS$,
    \item the data has a spike in the offensive end zone (where the offense would score a touchdown), which grows relative to the starting $LOS$, and
    \item the data has a spike in the defensive end zone (where the defense would score a touchdown), which grows inversely to the starting $LOS$.
\end{itemize}
In context, each of these features make sense. Turnovers require the ball to change possession from offense to defense, and thus are more likely to occur where the most interaction between opposing players occur, which is around the line of scrimmage. The spikes in each end zone are another example of the truncation effect occurring, where the ball is either returned for a touchdown by breaking free of potential tacklers originally on the offensive side of the ball, or by taking control of the ball in the end zone for a touchback.

However, the data is quite sparse with regards to plays with turnovers, particularly when it is split by starting $LOS$. Therefore, using historical data to accurately fit the previously observed data seems to be a unnecessary burden to achieve at best marginal precision levels. Thus, we will model the starting field position of the defense using a simple Gaussian distribution, centered at the original value of $LOS$. To deal with the truncation spikes, we will model each end point probability by using the CDF of the normal distribution evaluated at each of the points, using the regular CDF evaluation for the lower tail and one minus the CDF evaluation for the upper tail. This will allow for a point mass at the touchdown and touchback line of scrimmages. The touchback probability will then be added onto the touchback $LOS$, which is currently $LOS = 80$ (offense has ball on their own 20 yard line) for non-kickoff plays.

What remains is choosing the variances for the turnover probabilities, and adjusting the magnitude of the densities to match that of those observed. Via examination, we noticed that interceptions tend to have a much greater variance in return $LOS$ than that of fumbles, likely due to the distance between defenders and offenders upon the event taking place giving interceptions a higher chance of being returned for meaningful yardage than fumbles. Therefore, we assigned a standard deviation of 10 yards to fumbles and 25 yards to interceptions. 

Finally, using the numbers provided by Burke (2010), we will adjust the distributions to match the overall probability of their corresponding rates. Assuming that fumbles are equally distributed by $LOS$, fumbles are lost on 0.65\% of run plays, and 0.97\% of pass plays, and thus the sum of the probabilities of fumbles will equal those values on their respective plays. For interceptions, we observed a trend in interception rates that is related to the $LOS$, therefore we let the interception rate be dynamic and responsive to the base preplay information for pass plays. This value is computed using the data used to compute the posterior distributions for pass plays, including an interception as one of the probable options. From there, the return yards are scaled by this interception rate, and we obtain the full distribution for pass plays by adding in the fumble information and scaling the distribution accordingly. 

\subsection*{B.7 Dealing with Truncation}
A common theme with modeling each of these play types is the truncation effect that comes from the end zones. For example, if a team runs a pass play at $LOS = 70$, somewhere around 5 to 10 percent of the time, a ball carrier will ``break free" from the defense and pick up an extra 10 to 30 yards that was not originally designed to be a part of the play (although, it may have been conceived to be a potential opportunity that the play could create). Thus the play may go for 20-40 yards. When this play type is repeated several times, we see the distribution of play results spread out smoothly along the continuum of options, sometimes creating modes but otherwise fairly smooth. However, if this same play was run several times (assuming similar levels of success) from $LOS = 30$, when this ball carrier breaks free, the maximum amount of yards the player could obtain is now limited by the end zone. Thus, plays where the ball carrier may pick up a smooth distribution of yards between 20 and 40 yards is truncated, having a spike at the value of $LOS$ yards gained. This is observed often in the data, in particular for values of $LOS$ greater than 95 or less than 15.

Fortunately, no extra efforts must be taken to deal with this phenomenon. As described in previous subsections, for pass plays we have opted to separately mark the unit mass of a positive touchdown within the categorical distribution. For run plays, upon examination of the final posterior plots, we notice that the mixture model allows for these spikes in probability to be accounted for naturally within the distribution. This, any truncation effect is included in the modeling process, and no further adjustments will be made.

\subsection*{B.8 Dealing with Penalties}
Penalties play a very large role in the game of football, as an inopportune offensive penalty can set an offense back to near irrecoverable distances to pickup a first down, or a defensive pass interference call could set up an offense on the one yard line, almost guaranteeing a score. Penalties differ both in their frequency and their significance between run and pass plays, where run play penalties tend to penalize the offense (holding, illegal blocks, etc.), and pass play penalties tend to penalize the defense (pass interference, defensive holding, roughing the passer, etc.) Lopez (2014). With this in mind, it is certainly a topic that plays a meaningful role in any play calling model. However, due to data limitations, we have not included penalties within this system.

This is likely the largest omission within our methods. Unfortunately, due to the structure of the data, information about the intended play type was often masked when penalties occurred, usually being marked as ``no\_play" with the inclusion of the play's details within the description. Therefore, in order to incorporate information about penalties and their specific occurrence and influence of run and pass plays, a massive amount of data reformulation would have to occur. Therefore, penalties were excluded from the results of this research, acting under the incorrect assumption that penalties have a net zero effect on the probabilities of each play type. 

\subsection*{B.9 Time}
The component of time is uniquely used for helping determine the time remaining at the start of the subsequent play after choosing a specified action. The rules of the NFL dictate that different plays have different time distributions, and therefore must be modeled separately. Additionally, through careful analysis we have noticed that the distribution of time is also dependent upon the time remaining in the game and the score differential (at least in terms of which team is leading). In general, it breaks down as expected: in late game scenarios, teams with leads tend to run plays that drain more time, and vice versa for trailing teams, despite the actual play type. 

Regardless, the time draws are modeled using three pieces of information about the play: the play type, the score differential from the offense's perspective, and an indicator of whether or not there are more than five minutes left in the game. The time of five minutes remaining was specifically selected to correspond with the NFL's rules of when the clock should be stopped, which changes at the point of five minutes remaining. A fourth piece of information, a boolean indicating whether the play is a fourth down attempt, is used to handle the clock stoppage that occurs after a change of possession on fourth down plays (assuming a change of possession occurs). Using this information, full distributions of each time distribution could be made using the same mixture of normals model used for modeling run probabilities. This model was chosen as the distributions found typically displayed multimodal, yet smooth curves, that were well represented when fit using this method. 

From here, the distributions were used for each play in the late game scenarios, drawing several times from the corresponding distribution upon selecting a play type. This is opposed to using the full distribution to model the totality of future options as we did with the play outcomes. This decision was made to ease the computational burden, which would increase in orders of magnitude had the full distribution been used for each play. The error created by this method is minimized by reevaluating utility values for multiple draws of time and averaging the results.

It's also worth noting that some major time controlling features of the game were not included in this methodology, namely timeouts, kneel plays, and spike plays. Each coach is afforded three timeouts per half which they can use to stop the clock, allowing their team to preserve the time remaining to allow them more opportunities to run plays. Kneeling and spike the balls are opposite plays, with the former allowing a team to use one of their plays to run the clock without attempting to gain yards, and the latter allowing a team to stop the clock with no gain. While these are obviously key features of any late game strategy, implementation would have made it more difficult to discuss results in a broad sense, and thus is another opportunity for future research efforts.  

\end{document}